\pdfoutput=1
\documentclass{article} %
\usepackage{iclr2027_conference,times}

\usepackage{amsmath,amsfonts,bm}

\def\eqref#1{equation~\ref{#1}}

\def\1{\bm{1}}

\DeclareMathAlphabet{\mathsfit}{\encodingdefault}{\sfdefault}{m}{sl}
\SetMathAlphabet{\mathsfit}{bold}{\encodingdefault}{\sfdefault}{bx}{n}

\usepackage{hyperref}
\hypersetup{colorlinks=true, linkcolor=red, citecolor=blue, urlcolor=magenta}
\usepackage{url}

\usepackage{graphicx}
\usepackage{subcaption}
\usepackage{mathtools}
\usepackage{multirow}
\usepackage{pifont}
\usepackage{wrapfig}
\usepackage{tikz}
\usetikzlibrary{arrows.meta}
\usepackage{booktabs,multirow,graphicx,pifont,amsmath,xspace}

\usepackage{amsmath,amssymb}
\usepackage{algorithm}
\usepackage{algpseudocode}
\usepackage{makecell}
\newsavebox{\tblbox}
\newsavebox{\imgbox}

\definecolor{keptbg}{HTML}{D6F2E2}
\definecolor{keptfg}{HTML}{1B6B47}
\definecolor{heldbg}{HTML}{FBEAC6}
\definecolor{heldfg}{HTML}{8A5A12}

\newcommand{\concept}{$\mathcal{C}$\xspace}
\usepackage{tikz}
\usepackage{xcolor}
\usepackage{booktabs}
\usepackage{makecell}
 
\definecolor{barenc}{HTML}{2A78D6}   %
\definecolor{bardec}{HTML}{EB6834}   %
\definecolor{bardecl}{HTML}{F5B799}  %
\definecolor{bartrack}{HTML}{E8E7E1} %
\definecolor{bardash}{HTML}{898781}  %
 
\newlength{\barlen}
\title{
Concept Generalization in Visual Brain Decoding via Encoder-Generated fMRI

}

\author{
Peter Yongho Kim$^{1,3}$, Juhyeon Park$^{2,3}$, Dongsheng Li$^{3}$, \\
\textbf{Yansen Wang$^{3}$, Dongqi Han$^{3}$, Mingqing Xiao$^{3}$\thanks{Co-corresponding author.}\phantom{\textsuperscript{*}},
Taesup Moon$^{1,2,4}$}\footnotemark[1] \\
$^{1}$ECE, Seoul National University,
$^{2}$IPAI, Seoul National University, \\
$^{3}$Microsoft Research Asia,
$^{4}$ASRI / INMC / AIIS, Seoul National University\\
\texttt{\{peterkim98, parkjh9229, tsmoon\}@snu.ac.kr},\\
\texttt{\{dongsli, yansenwang, dongqihan, mingqingxiao\}@microsoft.com}
}

\iclrfinalcopy %
\begin{document}

\maketitle
\lhead{Preprint. Under review.} %
\begin{abstract}

Visual brain decoders are typically trained and evaluated on stimuli drawn from similar semantic distributions, leaving their ability to reconstruct unseen concepts poorly understood. We study this concept generalization problem by withholding semantic concepts from a target subject's training data. We find that three recent fMRI-to-image decoders degrade substantially for the held-out concepts, even when the concept remains available for multi-subject models. In contrast, brain encoders that map a pretrained visual representation directly to voxels are markedly more robust to the same semantic holdout. Motivated by this asymmetry, we use such an encoder to generate synthetic fMRI responses for new images and add the resulting image-fMRI pairs to decoder training. Across three decoders, four subjects, and twelve semantic concepts, synthetic data recovers a substantial fraction of the performance lost on unseen concepts without requiring additional, time-consuming fMRI acquisition. We further show that the same approach improves general decoding performance across a wide range of data budgets. 
Analyses of encoder training conditions, encoder designs, and synthetic-data controls show that the gains depend on both encoder generalization and the semantic content of the synthetic supervision, rather than merely increasing the amount of training data.
Together, our results show that encoder-generated fMRI can transfer semantic generalization to decoders.
\end{abstract}

\section{Introduction}

Recent fMRI-to-image decoders produce reconstructions that are often semantically faithful to the viewed stimulus \citep{mindeye2, brainit}. These results are obtained under a protocol in which training and test images are drawn from the same pool of semantic concepts: a decoder asked to reconstruct a zebra at test time has already been trained on other zebras and the responses they evoked. The protocol therefore leaves a basic question unanswered: \textit{Can a decoder reconstruct a concept it was never trained on?} A deployed decoder will inevitably face such concepts. \citet{shirakawa2025spurious} recently showed that the semantic overlap between training and test stimuli inflates apparent generalization, but the only remedy available so far is the one that limits brain decoding in the first place: collecting paired image-fMRI data for every new concept and every new subject, a time-consuming and expensive process.

\begin{figure}[t]
  \centering
  \subcaptionbox{Concept generalization setting, \concept{} = \emph{animal}.
                 \label{fig:setting}}
                [0.50\textwidth]
                {\includegraphics[width=\linewidth]{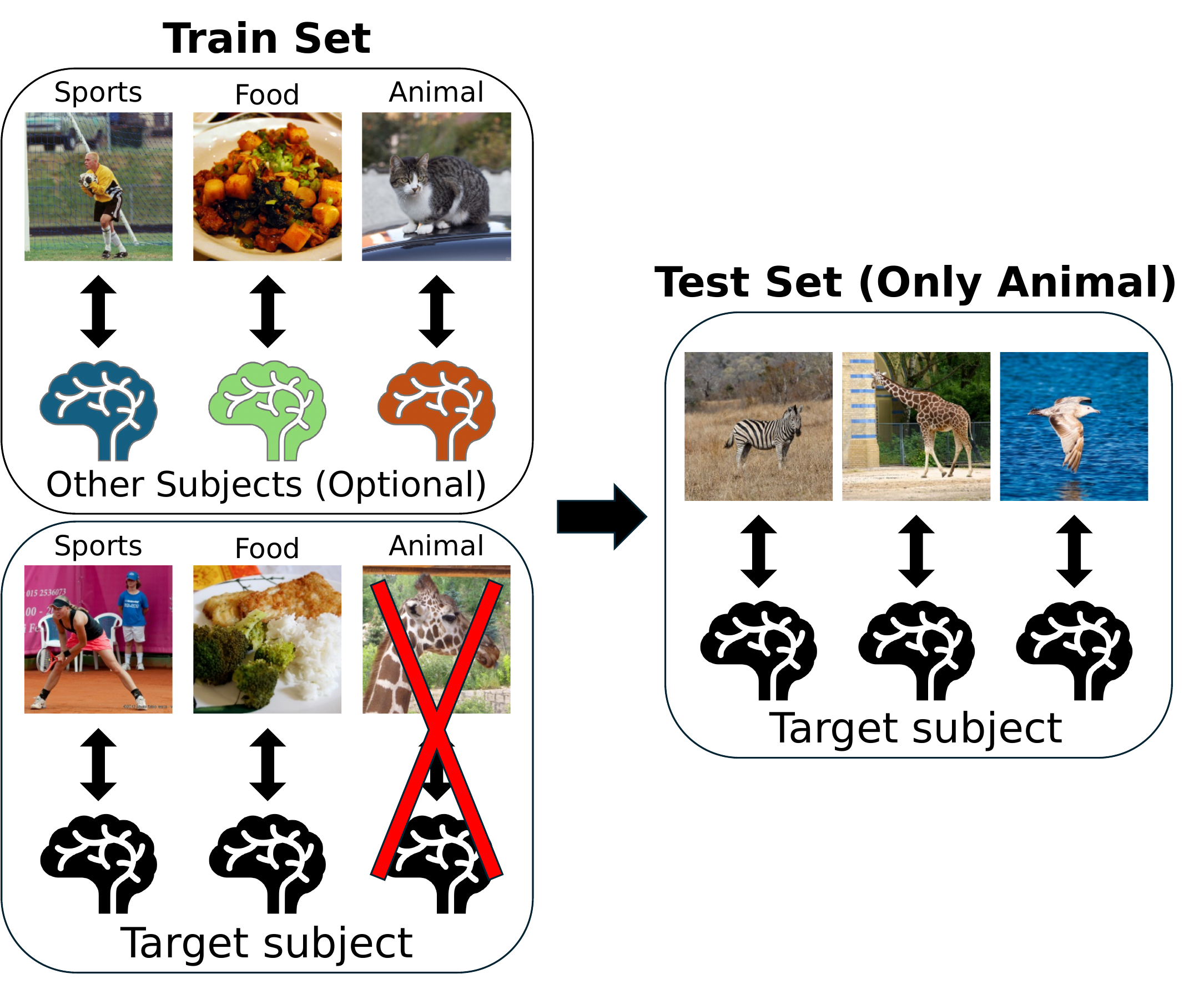}}
  \hfill
  \subcaptionbox{Recall retained when \concept{} is held out.
                 \label{fig:retained}}
                [0.45\textwidth]
                {\includegraphics[width=\linewidth]{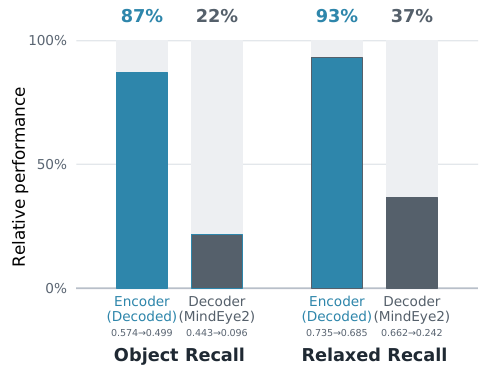}}
  \vspace{-2mm}
  \caption{\textbf{The concept generalization setting and the encoder-decoder asymmetry.}
  (a) Images containing \concept are removed only from the target subject's training set and not from other subjects; the model is evaluated only on images containing \concept.
  (b) The encoder retains most of its concept-specific recall under this holdout, while the MindEye2 decoder~\citep{mindeye2} loses most of it. Encoder recall is measured by decoding its predicted fMRI with a MindEye2 trained with \concept present and with additional synthetic data (See Sec.~\ref{sec:encoder-variants} for more details).
  }
  \label{fig:setting_and_gap}
  \vspace{-4mm}
\end{figure}

We make this question measurable with a \emph{concept generalization} setting (Figure~\ref{fig:setting}). We take a concept \concept as a COCO supercategory, remove every image containing \concept from the target subject's training set, and evaluate only on images containing \concept. Because standard reconstruction metrics compare global embeddings and can score a reconstruction highly even when the object of interest is absent, we complement them with object-detection metrics that ask whether \concept itself is recovered. Under this setting, three recent decoders \citep{ozcelik2023natural, mindeye2, brainit} degrade sharply across four subjects and all twelve supercategories; \textit{e.g.,} MindEye2~\citep{mindeye2} retains only 22\% of its concept recall. Notably, the degradation persists even though \concept remains available in the data of the other subjects used for cross-subject pre-training or joint training. Decoders thus largely fail to transfer a concept seen in other brains to a subject who never saw it.

Brain encoders, which map images to fMRI, behave differently under the same holdout: when their predicted fMRIs are decoded, an encoder trained without \concept from the target subject retains 87\% of its Object Recall, much higher than the case of the MindEye2 decoder (Figure~\ref{fig:retained}; Sec.~\ref{sec:motivation}).
Although these two ratios compare two specific pipelines rather than isolate an intrinsic advantage of the encoding direction, the asymmetry suggests using the encoder to transfer its robustness to the decoder through data. We generate synthetic fMRI responses for images containing \concept and add the resulting image-fMRI pairs to the decoder's training set. Because the intervention acts on the data alone, it applies to any decoder architecture or training strategy without modification, and it requires no additional scanning. Across the three decoders, the synthetic data recovers a large fraction of the generalization gap without any real fMRI for \concept.

Since the encoder is trained on the same data as the decoder, why should its synthetic data help? By varying the encoder's training conditions and the composition of the synthetic data, we identify two sources of the gain. First, the encoder is inherently robust to unseen concepts: even an encoder that never saw \concept in any subject yields substantial gains, which we trace to its pretrained visual backbone: a ridge regression on frozen backbone features already helps and a different pretrained backbone gives similar gains, while a randomly initialized one gives almost none. This robustness is not automatic across encoder designs, as synthetic data from SynBrain~\citep{mai2026synbrain} gives little recovery. Second, the encoder transfers exposure to \concept from other subjects, which decoders fail to exploit on their own. Controls further show that the gain is not a data-volume effect; it requires correctly paired synthetic responses to images of \concept itself, although the synthesis pool need not be selected for \concept as long as it covers it.

Finally, the use of encoder-generated synthetic data is not specific to concept generalization. When the synthesis pool consists of images unseen by the target subject, it improves all three decoders at every target-subject data budget from 225 to 9,000 images for most metrics, by an amount equivalent to 1.9--2.4$\times$ as many real training images, depending on the decoder (Sec.~\ref{sec:scaling-synthetic}).

While using encoder-generated fMRI to train decoders is not new in itself (Sec.~\ref{sec:related-synthetic}), our contribution is a systematic study of concept generalization in recent decoders and of the conditions under which synthetic data addresses it.
Our contributions are as follows. (i) We introduce the concept generalization setting, with concept-level recall metrics, and show that three recent decoders fail it even under cross-subject training. 
(ii) We show that brain encoders built on a pretrained visual representation are far more robust to the same holdout, and that transferring this robustness through encoder-generated fMRI for images of the held-out concept recovers a large fraction of the gap across three decoders, four subjects, and twelve concepts.
(iii) We show that the gain stems from the encoder's inherent robustness and from cross-subject transfer rather than from added data volume, and identify what the synthetic data must contain (the held-out concept, correctly paired) and what the encoder must provide (a pretrained visual representation with a direct readout to voxels). (iv) We show that the same recipe improves decoding at various data budgets on most metrics, equivalent to 1.9--2.4$\times$ as many real training images.

\section{Background and Related Work}
\subsection{Visual Brain Decoding and Encoding}
{

}

From neuroimaging experiments that produce paired visual stimulus-fMRI response data, two complementary tasks can be defined: brain decoding and brain encoding. 
\emph{Brain decoding} learns a map from fMRI to image, reconstructing the stimulus given the brain response.
\emph{Brain encoding} inversely learns a map from image to fMRI, predicting the brain response given the stimulus. One of the most widely used datasets is the Natural Scenes Dataset (NSD)~\citep{nsd}, where eight subjects viewed images drawn from COCO~\citep{coco}. NSD is the main dataset we use in this work. 

By incorporating diffusion models, brain decoding models rapidly evolved, albeit requiring separately trained models for each subject \citep{ozcelik2023natural, mindeye1}.
Since this paired data is costly, some recent decoders utilize cross-subject (pre-)training to help increase the performance for other subjects and therefore reduce the per-subject data burden \citep{mindeye2, wang2024mindbridge,huo2024neuropictor,brainit}.

On the encoding side, the Universal Brain Encoder
\citep{beliy2024wisdom} learns voxel embeddings that attend to DINOv2 \citep{dinov2} features with LoRA~\citep{hu2022lora} tuning. The parameters are shared across voxels and subjects, which makes it applicable to different voxel counts and transferable between multiple subjects with minimal fine-tuning, and we use it as our brain encoding model for the experiments. 
Other recent encoders include SynBrain \citep{mai2026synbrain}, which models the response probabilistically, and TRIBE v2 \citep{tribev2}, a multimodal encoder which is also capable of predicting subject-agnostic cortical surface responses. 

\subsection{Semantic Generalization for Visual Brain Decoding}
Whether a decoder can handle concepts absent from its training set is a question that has been raised before~\citep{horikawa2017generic}, but one that most recent decoding work does not engage with.
Since NSD draws training and test images from the same pool, \citet{shirakawa2025spurious} argue that the resulting overlap inflates performance. They cluster NSD stimuli in CLIP text-feature space, withhold one cluster from training, and find that decoder performance drops and that the standard metrics overstate their generalization capabilities. 
Their proposed solutions concern dataset design and representation choice rather than a method that closes the gap for an existing dataset.

We later propose a \emph{concept generalization} setting that builds on this diagnosis. We define the held-out concept from ground-truth annotations and evaluate decoders that train on multiple subjects (Sec.~\ref{sec:concept_generalization}), use additional metrics to measure whether the concept itself is recovered (Sec.~\ref{sec:setting}), and, unlike prior work, show a working solution that does not need to collect additional fMRI (Sec.~\ref{sec:encoder_synthetic}).

\subsection{Using Encoder-generated Synthetic fMRI for Decoding}
\label{sec:related-synthetic}
To reduce the burden of data collection, a line of work produces synthetic neural responses for images that were never shown to the subject.
Earlier work trains the decoder on such images through a fixed encoder, which supplies surrogate responses \citep{st2018generative} or a cycle-consistency signal \citep{gaziv2022self}, the latter reporting gains on categories never seen in training.
Their unpaired image pool is, however, not chosen with the test categories in mind: in \citet{gaziv2022self} the test categories are absent from the paired data and from the unpaired pool alike.
More recently, brain encoders have been used to generate synthetic fMRI for decoder training: Brain-IT \citep{brainit} and SynBrain \citep{mai2026synbrain} add encoder-predicted responses to images drawn from the same distribution as the real stimuli, in full-data and few-shot settings, and \citet{benchetrit2026boosting} study synthetic cortical surface maps generated by TRIBE v2, focusing on common, subject-agnostic synthetic data and leaving subject-specific synthesis as an open question.
None of these works examines whether the synthetic pool covers the concept the decoder must generalize to, nor evaluates on held-out concepts, which is the gap we address.

\section{Data-limited Settings for Brain Decoding}

As paired image-fMRI training data is slow and expensive to acquire,  the \textbf{data efficiency} of brain decoders is essential to gauging their practical utility.
We use the term data efficiency to represent the ability to attain satisfactory performance, even when the training data is limited for the target subject (the subject it will be evaluated on).
For decoders that train with multiple subjects~\citep{mindeye2, brainit}, a common practice is to only limit the data from the target subject, not the other subjects.
The reasoning is that in practice, data from other subjects is a one-time cost shared by every deployed decoder, whereas target-subject data must be collected for each subject. 

Target-subject data can be restricted along two distinct axes. The commonly addressed one is quantity of the available data. The other is semantic coverage: which concepts those data depict.
We introduce the semantic axis as the concept generalization setting (Sec.~\ref{sec:concept_generalization}), and the quantity axis as the limited-data decoding setting (Sec.~\ref{sec:few-shot-generalization-setting}). 

\subsection{Concept Generalization: Limiting the Semantic Coverage}
\label{sec:concept_generalization}

The concept generalization setting (Figure~\ref{fig:setting}) measures how a decoder behaves on a concept for which the target subject contributed no training data.
Given a target concept \concept, a concept which we want to measure the generalization capability for, we exclude every image containing \concept from the training set of the target subject. The model is then evaluated only on images containing \concept.

\paragraph{Defining the concept.}
To introduce a large semantic gap between train and test time, we take \concept to be a COCO supercategory, for example \emph{animal}. We filter images using their ground-truth COCO annotations, where an image is considered to contain \concept if its annotation includes at least one instance of any subcategory belonging to \concept. 
We use ground-truth annotations rather than clusters in a learned feature space \citep{shirakawa2025spurious} for two reasons. A cluster-based holdout retains training images in which the concept is present but not dominant, which leaks the concept into training; the annotation rule removes it entirely. It also keeps the holdout independent of any model's feature space, so that the holdout and the evaluation are not defined by the same representation.
We use all 12 supercategories, each defining its own training set and  separately trained encoders and decoders. 

\paragraph{Can concept exposure from other subjects solve concept generalization?}
Given that recent decoding and encoding methods optionally allow multi-subject training, whether \concept is included for the other subjects must also be accounted for. 
For the main experiments we assume \concept remains available for the subjects used in pre-training, which we call the \textbf{transfer-concept setting}.
For multi-subject models, this gives an opportunity to learn about \concept from other brains, a question that per-subject holdout protocols \citep{shirakawa2025spurious} cannot pose.
It also makes the setting deliberately favorable to existing models: if they still lose performance, concept generalization cannot be achieved by their inherent cross-subject transfer capabilities and a different solution is needed.
We additionally explore settings where \concept is also absent for other subjects, as part of Sec. \ref{sec:encoder-variants}.

\subsection{Limited-data Decoding: Limiting the Quantity}
\label{sec:few-shot-generalization-setting}

Most existing work on data efficiency instead limits the amount of data: the target subject's training set is subsampled to a small budget while the pre-training subjects retain their full data. 
We follow the same protocol and specify the exact budgets in Sec. 5.

We note that the two settings are not entirely independent. Subsampling a training set may under-represent or even exclude some of the semantic content. For example, for our random 225-image training splits, equivalent to the \emph{1 hour} splits from previous work~\citep{mindeye2, brainit}, on average 31.5 out of the 80 COCO categories appear on 5 or fewer images from the training set. The more aggressively the training data is limited, the more of the concept space will be effectively unseen, and the more a few-shot problem also turns into a concept generalization problem.

\begin{figure}[t]
  \centering
  \includegraphics[width=.9\linewidth]{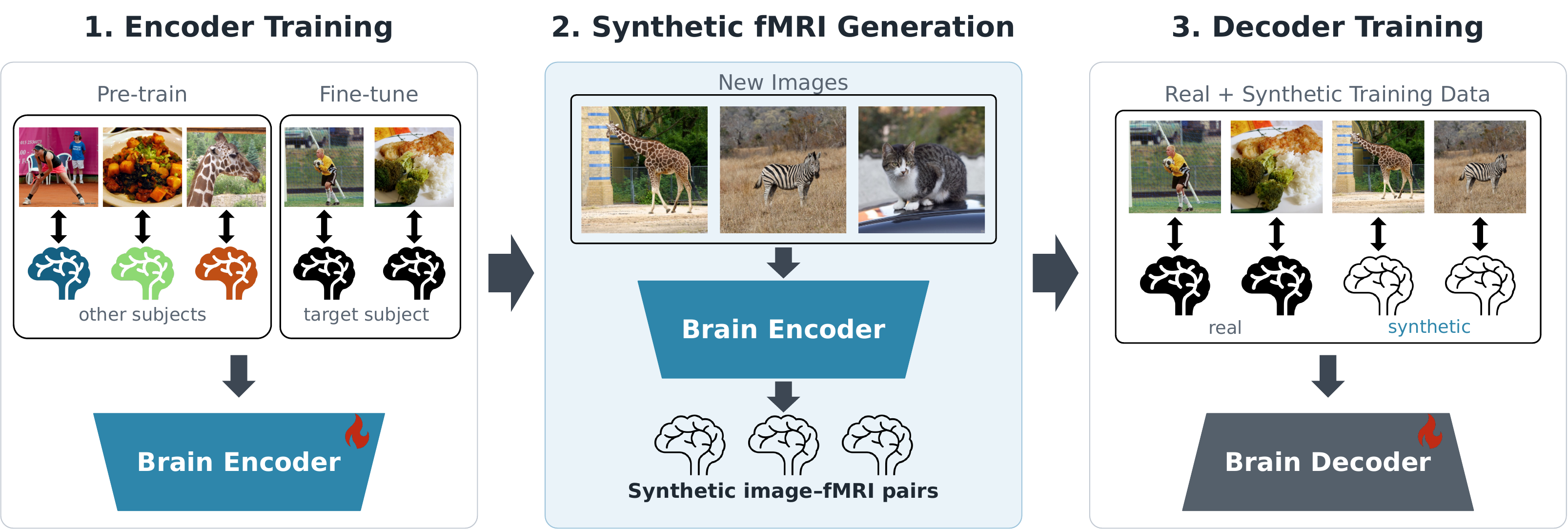}
  \caption{\textbf{Encoder-generated synthetic fMRI for decoder training.}
  The encoder is first pre-trained and fine-tuned to the target subject (step~1), applied to a new pool of images to generate synthetic fMRI (step~2), and the synthetic pairs are added to the decoder's training set (step~3).}
  \label{fig:synthetic_explain}
\vspace{-4mm}
\end{figure}

\section{Generating Synthetic Training Data via Brain Encoders}
\label{sec:encoder_synthetic}

\subsection{motivation for using brain encoder}
\label{sec:motivation}

Using encoder-generated synthetic data for concept generalization is motivated by the asymmetry illustrated in Figure~\ref{fig:retained}. To compare the encoder and decoder on similar footing, we score the encoder by \emph{decoded recall} (defined in Sec.~\ref{sec:encoder-variants}): its predicted fMRI for images with \concept is reconstructed by a fixed MindEye2 readout trained with \concept present, so that the readout's own concept-generalization limits do not confound the measurement, and scored with the concept metrics given in Sec.~\ref{sec:setting}. Under this holdout, the encoder retains 87\% of Object Recall and 93\% of Relaxed Recall, whereas MindEye2 retains 22\% and 37\%, respectively. 
Since the encoder and the decoder convert information in opposite directions, these two ratios describe two specific pipelines and do not isolate an intrinsic advantage of the encoding direction; we use them as motivation, and test the actual utility of the encoder's predictions by training decoders on them (Sec.~\ref{sec:experiments}). The pattern persists with a readout trained on real data only (Appendix~\ref{app:decoded-recall}).
In this comparison, the encoder is more robust than the decoder, which motivates using its predictions to train decoders; Sec.~\ref{sec:encoder-variants} further examines why.

\subsection{Recipe for Generating Synthetic Training Data}
\label{sec:synthetic_recipe}

The cost of paired image-fMRI data lies in acquiring the fMRI, not the image. A brain encoder can therefore supply the missing half by predicting responses to novel images, and the resulting pairs can be used to train brain decoders. The procedure is summarized in Figure~\ref{fig:synthetic_explain}.

\paragraph{The recipe.} Let $\mathcal{D} = \{(x_i, y_i)\}_{i=1}^{N}$ denote the target subject's training set, where $x_i$ is an image and $y_i$ the corresponding measured fMRI response. We train a brain encoder $E_\phi : x \mapsto \hat{y}$ on $\mathcal{D}$, initializing it from an encoder pre-trained on other subjects, so the encoder sees no target-subject fMRI beyond what the decoder already has. Given a new pool of images $\mathcal{U} = \{u_j\}$, we form a synthetic training set $\mathcal{S} = \{(u_j, E_\phi(u_j)) : u_j \in \mathcal{U}\}$ and train the decoder on $\mathcal{D} \cup \mathcal{S}$. 
The choice of $\mathcal{U}$ is the one component that varies across our experiments, and we specify it for each setting in Sec.~\ref{sec:experiments}. Because this is a data-level intervention, the synthetic pairs can train any decoding architecture without modification, regardless of its input shape or training strategy.

\section{Experimental Results}
\label{sec:experiments}

\subsection{Experimental Setting}
\label{sec:setting}

All experiments are conducted on the Natural Scenes Dataset (NSD)~\citep{nsd}.
Following common practice, we experiment with four subjects (1, 2, 5 and 7) who completed all scanning sessions, and report the averaged result.
We experiment with three image decoders: Brain-Diffuser~\citep{ozcelik2023natural}, MindEye2~\citep{mindeye2}, and Brain-IT~\citep{brainit}. Decoders are trained with multiple subjects if possible: Brain-Diffuser can only be trained with a single subject, MindEye2 is first pre-trained with other subjects and then fine-tuned, and Brain-IT is jointly trained with all subjects. Refer to Appendix~\ref{app:additional_exp_details} for more details.

Note that although Brain-IT includes external synthetic data training in the original paper, similar in spirit to our synthetic data training, we consider this process to be a training technique independent of the core Brain-IT architecture. To keep the training process fair and consistent with other decoding models, and to measure the precise effect of synthetic data, we use the base Brain-IT model without this external data augmentation, just like the other models.

We use the Universal Brain Encoder ~\citep{beliy2024wisdom} as our synthetic-data generator. The encoder is pre-trained on other subjects and adapted to the target subject, using the same training data.

\paragraph{Evaluation.}
We report the standard decoding metrics: low-level (Pixelwise correlation, SSIM, Alex(2), Alex(5)) and high-level (Inception, CLIP, EffNet and SwAV).
For concept generalization experiments, we additionally report \emph{Object Recall} and \emph{Relaxed Object Recall}, adapted from SEED~\citep{seed}: using object detectors, we measure the ratio of reconstructions that exactly contain the object (Recall) or at least contain the same supercategory (Relaxed Recall), limited to objects that belong to \concept. For example, if \concept is \emph{animal}, and the ground truth includes a zebra, Recall counts the reconstruction correct if it also includes a zebra, while Relaxed Recall counts it correct if it contains any animal. Identification metrics compare global embeddings against a distractor and can remain above chance even when the held-out concept is absent from the reconstruction; \citet{shirakawa2025spurious} report exactly this discrepancy under a semantic holdout. We therefore aim to measure concept recovery directly through these metrics. Refer to Appendix~\ref{app:metrics} for more details.

\begin{table}[t]
\centering
\caption{\textbf{Concept generalization for decoders.} Decoders are trained with/without \concept for the target subject and evaluated on images containing \concept. Adding synthetic data improves generalization towards \concept. Bold marks the better of the two unseen variants, with or without synthetic data.}
\label{tab:unseen-concept-decoder}
\small
\setlength{\tabcolsep}{5pt}
\begin{tabular}{llcccccc}
\toprule
& & \multicolumn{4}{c}{High-level} & \multicolumn{2}{c}{Object Detection} \\
\cmidrule(lr){3-6}\cmidrule(lr){7-8}
Decoder & Concept \concept & Incep $\uparrow$ & CLIP $\uparrow$ & EffNet $\downarrow$ & SwAV $\downarrow$ & Recall $\uparrow$ & Relaxed $\uparrow$ \\
\midrule
\multirow{3}{*}[-2pt]{Brain-Diffuser} & Seen & 90.2\% & 91.7\% & 0.761 & 0.397 & 0.330 & 0.584 \\
\cmidrule(lr){2-8}
& Unseen & 81.2\% & 82.5\% & 0.847 & 0.458 & 0.073 & 0.189 \\
& Unseen + syn. & \textbf{88.7\%} & \textbf{90.5\%} & \textbf{0.781} & \textbf{0.408} & \textbf{0.273} & \textbf{0.532} \\
\midrule
\multirow{3}{*}[-2pt]{MindEye2} & Seen & 95.9\% & 93.0\% & 0.636 & 0.340 & 0.443 & 0.662 \\
\cmidrule(lr){2-8}
& Unseen & 88.5\% & 83.5\% & 0.767 & 0.410 & 0.096 & 0.242 \\
& Unseen + syn. & \textbf{93.4\%} & \textbf{89.0\%} & \textbf{0.697} & \textbf{0.371} & \textbf{0.288} & \textbf{0.482} \\
\midrule
\multirow{3}{*}[-2pt]{Brain-IT} & Seen & 97.3\% & 95.0\% & 0.600 & 0.331 & 0.495 & 0.672 \\
\cmidrule(lr){2-8}
& Unseen & 94.9\% & 91.9\% & 0.659 & 0.360 & 0.312 & 0.490 \\
& Unseen + syn. & \textbf{96.0\%} & \textbf{93.5\%} & \textbf{0.637} & \textbf{0.349} & \textbf{0.386} & \textbf{0.565} \\
\bottomrule
\end{tabular}
\vspace{-4mm}
\end{table}

\newlength{\reconw}
\setlength{\reconw}{0.10\linewidth}
\newcommand{\reconcell}[1]{%
  {\setlength{\fboxsep}{0pt}%
   \framebox[\reconw]{\parbox[c][\reconw][c]{\reconw}{\centering\tiny #1}}}}
\newcommand{\reconbox}[1]{%
  \parbox[c][\reconw][c]{\reconw}{\centering #1}}

\begin{figure}[t]
\centering
\setlength{\tabcolsep}{1.5pt}
\renewcommand{\arraystretch}{1.0}
\small
\begin{tabular}{@{}l@{\hspace{4pt}}c@{\hspace{4pt}}cc@{\hspace{3pt}}cc@{\hspace{3pt}}cc@{}}
\toprule
\multicolumn{1}{@{}c@{\hspace{4pt}}}{\multirow{2}{*}[-2pt]{\concept}}
     & \multirow{2}{*}[-2pt]{Ground Truth}
                    & \multicolumn{2}{c}{Brain-Diffuser}
                    & \multicolumn{2}{c}{MindEye2}
                    & \multicolumn{2}{c}{Brain-IT} \\
\cmidrule(lr){3-4}\cmidrule(lr){5-6}\cmidrule(lr){7-8}
     &
     & Unseen & Unseen + syn.
     & Unseen & Unseen + syn.
     & Unseen & Unseen + syn. \\
\midrule
Animal &
  \reconbox{\includegraphics[width=\reconw]{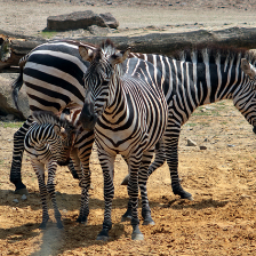}} &
  \reconbox{\includegraphics[width=\reconw]{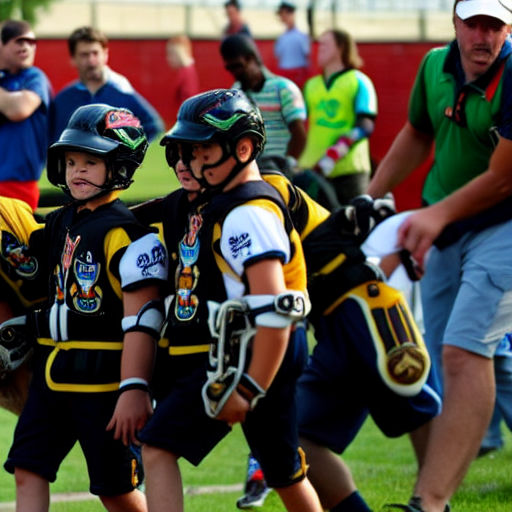}} &
  \reconbox{\includegraphics[width=\reconw]{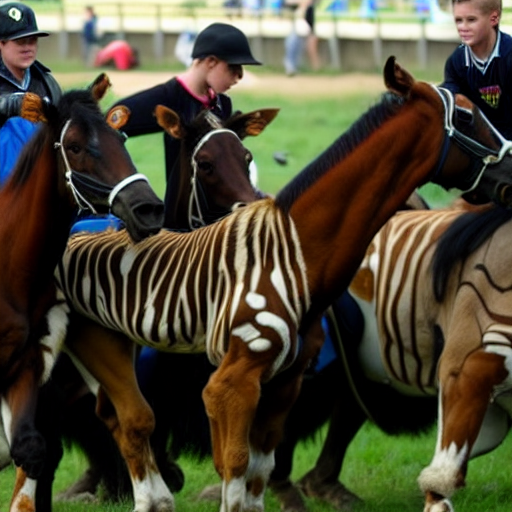}} &
  \reconbox{\includegraphics[width=\reconw]{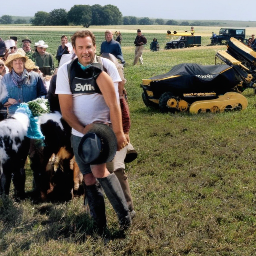}} &
  \reconbox{\includegraphics[width=\reconw]{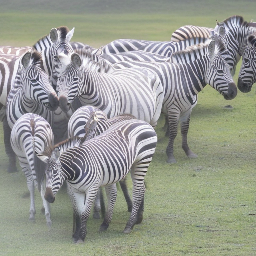}} &
  \reconbox{\includegraphics[width=\reconw]{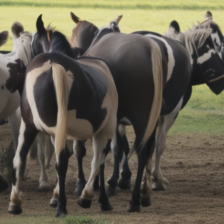}} &
  \reconbox{\includegraphics[width=\reconw]{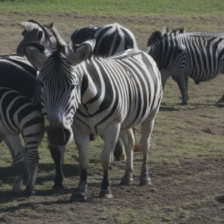}} \\[18pt]
Person &
  \reconbox{\includegraphics[width=\reconw]{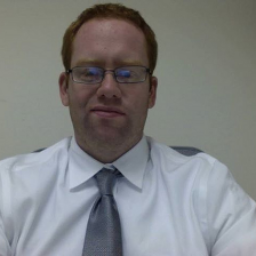}} &
  \reconbox{\includegraphics[width=\reconw]{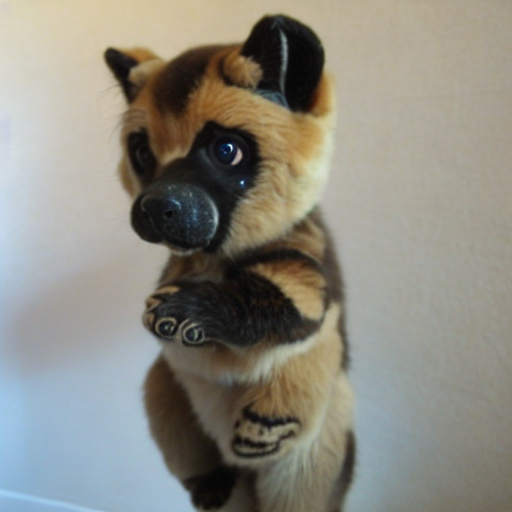}} &
  \reconbox{\includegraphics[width=\reconw]{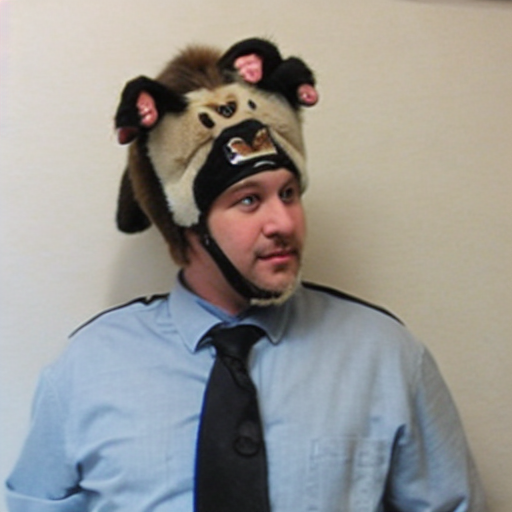}} &
  \reconbox{\includegraphics[width=\reconw]{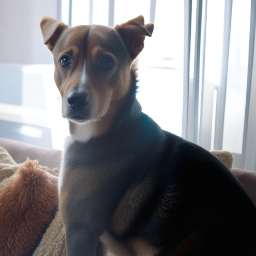}} &
  \reconbox{\includegraphics[width=\reconw]{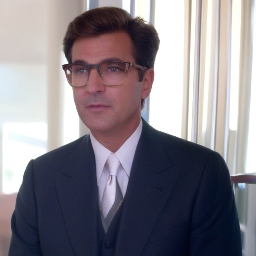}} &
  \reconbox{\includegraphics[width=\reconw]{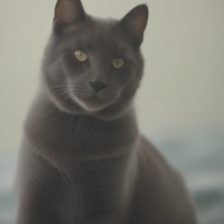}} &
  \reconbox{\includegraphics[width=\reconw]{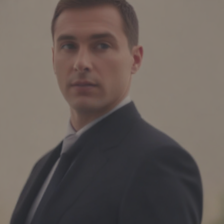}} \\[18pt]
Vehicle &
  \reconbox{\includegraphics[width=\reconw]{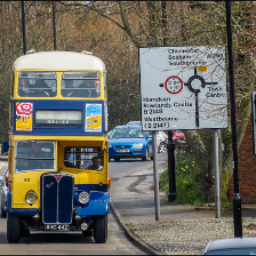}} &
  \reconbox{\includegraphics[width=\reconw]{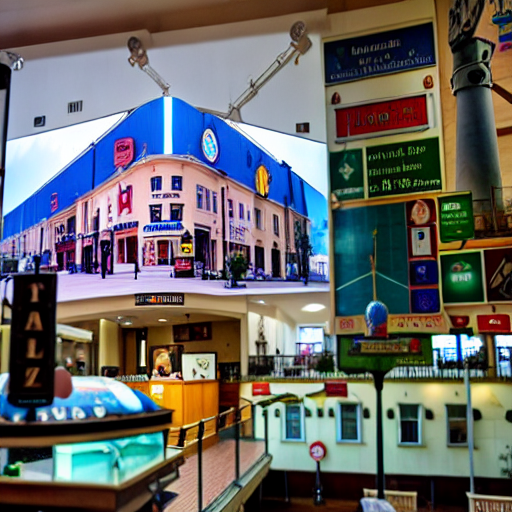}} &
  \reconbox{\includegraphics[width=\reconw]{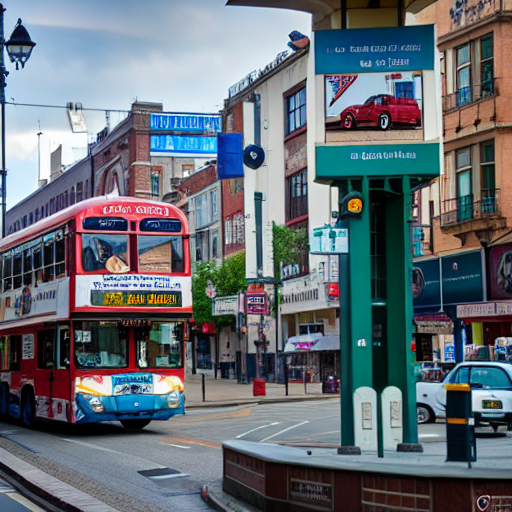}} &
  \reconbox{\includegraphics[width=\reconw]{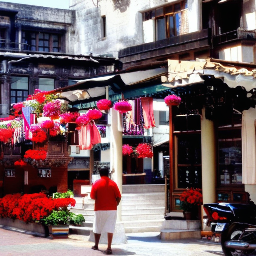}} &
  \reconbox{\includegraphics[width=\reconw]{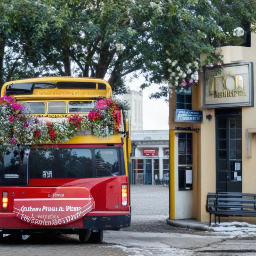}} &
  \reconbox{\includegraphics[width=\reconw]{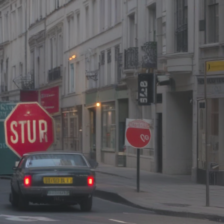}} &
  \reconbox{\includegraphics[width=\reconw]{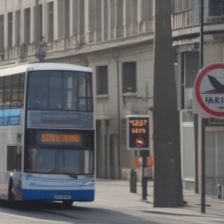}} \\
\bottomrule
\end{tabular}
\caption{\textbf{Qualitative reconstruction highlights.} We show the ground truth image and the corresponding reconstructions for each decoder, before or after adding synthetic data to training. While the baseline model fails to accurately reconstruct \concept, the synthetic variant is able to recover it.}
\label{fig:recon-holdout}
\vspace{-4mm}
\end{figure}

\subsection{Using Synthetic Data for Concept Generalization}
\label{sec:concept-generalization-results}

Table~\ref{tab:unseen-concept-decoder} reports the performance of each decoder depending on whether \concept was included in the target subject's training data and whether synthetic data was added. 
To generate synthetic fMRI, we use the images that are removed from the training set, to directly match the baseline condition (unseen) and the synthetic condition (unseen + syn.); Sec.~\ref{sec:encoder-variants} shows that a general pool not selected for \concept recovers nearly as much.
Low-level metrics, which the semantic holdout affects far less, are reported in Appendix~\ref{app:holdout-lowlevel}.

All three decoders show a concept generalization gap, largest on the recall metrics. This includes MindEye2 and Brain-IT, which were trained with \concept available from other subjects; as anticipated in Sec.~\ref{sec:concept_generalization}, cross-subject training alone does not transfer \concept to the target subject.
The gap is specific to \concept: on test images without \concept, the same models differ by an order of magnitude less (Appendix~\ref{app:complementary_results}), and removing an equal number of randomly chosen images without \concept from training costs essentially nothing on images of \concept (Appendix~\ref{app:random-deletion}).

When synthetic data is supplied, it closes a substantial portion of the gap. Weaker decoders recover more of the lost performance: Brain-Diffuser recovers 76-87\%, MindEye2 recovers 53-66\%, and Brain-IT recovers 37-53\% of the lost performance across high-level and recall metrics.

Figure~\ref{fig:recon-holdout} shows a few qualitative examples on how the baseline decoder fails to reconstruct the object belonging to \concept and how the synthetic-trained decoder is able to reconstruct the object.

\subsection{Using Synthetic Data Across Data Scales}
\label{sec:scaling-synthetic}

\begin{figure}[t]
  \centering
  \includegraphics[width=.95\linewidth]{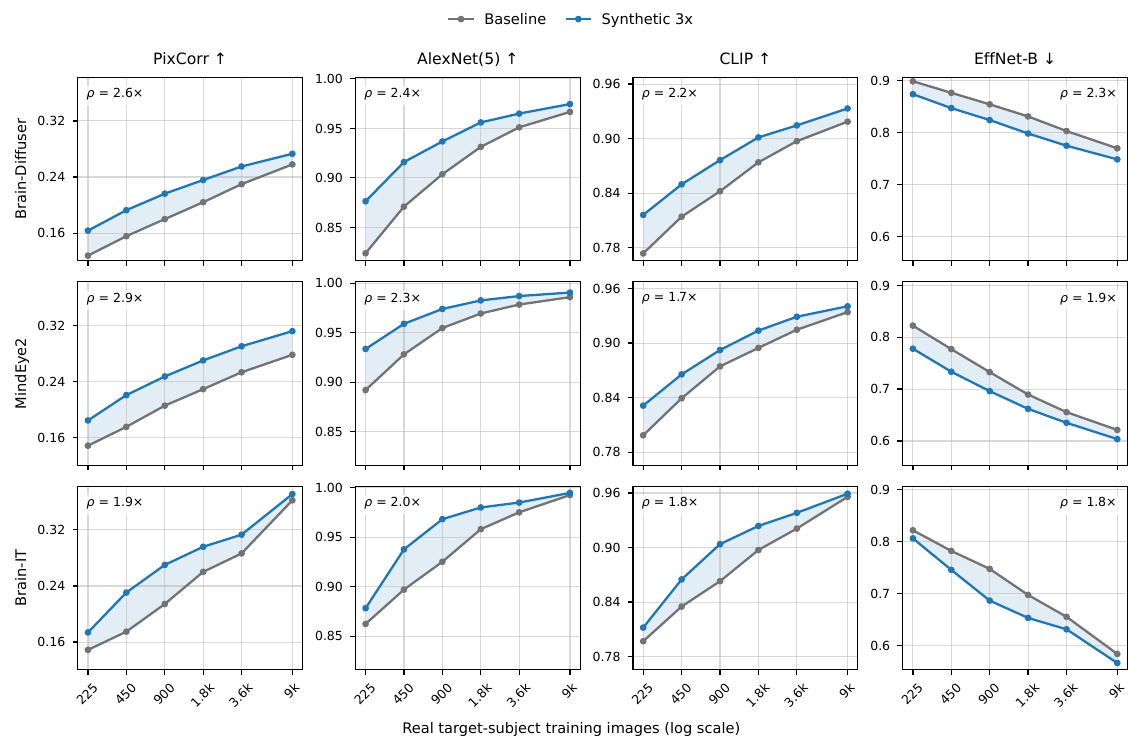}
\vspace{-4mm}
  \caption{\textbf{Effect of synthetic data across data scales}. The $x$ axis is in log scale. $\rho$ denotes the fitted data-efficiency gain. We show 4 metrics spanning the metric groups; the rest are in Appendix~\ref{app:few-shot-additional}.}
\vspace{-4mm}
  \label{fig:various_data_scale}
\end{figure}

We now turn to the quantity axis of Sec.~\ref{sec:few-shot-generalization-setting}, training decoders on
$x \in \{225, 450, 900, 1800, 3600, 9000\}$ real target-subject images, with and without
additional synthetic data. The synthesis pool $\mathcal{U}$ consists of NSD images shown to other
subjects but never to the target subject, with $|\mathcal{U}| = 3x$ so that real and
synthetic pairs contribute equal training volume: each real image carries three trials,
while the encoder produces one deterministic response per image. 
Appendix~\ref{app:scaling-synthetic} varies this ratio between real and synthetic data.

Synthetic data outperforms the baseline at every budget on every metric in Figure~\ref{fig:various_data_scale}, and both curves keep rising up to the full 9,000-image budget, indicating that the decoders are not yet saturated by target-subject data. The remaining metrics behave the same way (Appendix~\ref{app:few-shot-additional}), with the exception of SSIM, which varies little with budget and synthetic data does not consistently help it.

\paragraph{Quantifying the data-efficiency gain.}
The benefit can equivalently be read as the real data collection it saves. For each metric
we fit the single factor $\rho$ that aligns the baseline curve with the synthetic one,
$\hat{\rho} = \arg\min_{\rho} \sum_{x} \bigl( s(x) - \hat{b}(\rho\,x) \bigr)^{2}$, where
$b(x)$ and $s(x)$ are the baseline and synthetic scores at $x$ real target-subject images and
$\hat{b}$ is a monotone interpolant of the six measured budgets, held at its 9,000-image value beyond the measured range. 
The resulting $\hat\rho$, annotated in each panel of Figure~\ref{fig:various_data_scale}, reads directly as the amount of real data the
synthetic data is worth: $\hat\rho = 2$ means the same performance
would otherwise have required twice as many real training images of the target
subject. Since $\hat{b}$ is held constant beyond 9{,}000 images, $\hat\rho$ therefore summarizes the gain only within the measured range. Across the four metrics in Figure~\ref{fig:various_data_scale} and all
three decoders, $\hat\rho$ ranges from 1.7 to 2.9 (the remaining
metrics fall in the same range; confidence intervals in
Appendix~\ref{app:stats_scaling}), with a median over the seven identifiable metrics of 2.4 for Brain-Diffuser, 2.2 for MindEye2, and 1.9 for Brain-IT.

\paragraph{Extension to video decoding.}
As a preliminary check of whether the recipe transfers beyond static images, we apply it to fMRI-to-video decoding with NEURONS~\citep{wang2025neurons} on the cc2017 dataset~\citep{wen2018neural}, training on 10\% or the full amount of data.
The same Universal Encoder, jointly pre-trained on NSD and the other cc2017 subjects, predicts a response from a single frame of each clip, a framewise approximation that does not model temporal dynamics (Appendix~\ref{app:video}). 
Synthetic data improves the video decoder at both budgets, with the largest absolute gains on the semantic metrics: at the 10\% budget, video-level 50-way accuracy rises from 0.185 to 0.235, recovering 68\% of the gap to the full-data model, and at the full budget the gains remain significant on most metrics (Appendix~\ref{app:video}). That a still-image encoder improves a video decoder suggests the recipe can apply beyond image decoding, even under this approximation.

\section{Why does Encoder-generated Data Help?}
\label{sec:encoder-variants}

\begin{table}[t]
\centering
\caption{\textbf{Encoder variants as synthetic-data sources (supercategory generalization).}
Variants differ in whether \concept was available during cross-subject pre-training and target-subject adaptation (Target). \emph{Probe decoder}: a from-scratch MindEye2 trained on real data plus that encoder's synthetic pairs, evaluated on images containing \concept; the two oracle variants see real responses to \concept and cannot be used for synthesis. \emph{Encoder}: Pearson $r$ of the predicted fMRI, and decoded recall / decoded relaxed recall (D.Rec.\,/\,D.Rel.) of images reconstructed from it.}
\vspace{-2mm}
\label{tab:encoder-variants}
\footnotesize
\setlength{\tabcolsep}{2.1pt}
\begin{tabular}{lcc|cccccc|ccc}
\toprule
& \multicolumn{2}{c|}{\concept{} incl.\ for} & \multicolumn{6}{c|}{Probe decoder trained with synthetic data} & \multicolumn{3}{c}{Encoder} \\
\cmidrule(lr){2-3}\cmidrule(lr){4-9}\cmidrule(lr){10-12}
Encoder variant & Other & Target & Incep $\uparrow$ & CLIP $\uparrow$ & EffNet $\downarrow$ & SwAV $\downarrow$ & Rec.\ $\uparrow$ & Rel.\ $\uparrow$ & $r$ $\uparrow$ & D.Rec.\ $\uparrow$ & D.Rel.\ $\uparrow$ \\
\midrule
Scratch-oracle   & --        & \ding{51} & \multicolumn{6}{c|}{\multirow{2}{*}{\textit{n/a: trained on real fMRI for \concept}}} & 0.303 & 0.497 & 0.686 \\
Transfer-oracle  & \ding{51} & \ding{51} & \multicolumn{6}{c|}{}                                                                  & 0.350 & 0.574 & 0.735 \\
\midrule
Scratch-blind    & --        & \ding{55} & 91.4\% & 86.9\% & 0.729 & 0.390 & 0.218 & 0.427 & 0.277 & 0.290 & 0.492 \\
Transfer-blind   & \ding{55} & \ding{55} & 91.4\% & 87.2\% & 0.723 & 0.387 & 0.229 & 0.441 & 0.337 & 0.379 & 0.585 \\
Transfer-concept & \ding{51} & \ding{55} & 92.8\% & 88.5\% & 0.709 & 0.381 & 0.279 & 0.486 & 0.344 & 0.499 & 0.685 \\
\midrule
No synthetic               & \multicolumn{2}{c|}{--} & 87.2\% & 82.5\% & 0.777 & 0.418 & 0.087 & 0.228 & \multicolumn{3}{c}{\multirow{2}{*}{\textit{n/a: no encoder}}} \\
Real fMRI for \concept  & \multicolumn{2}{c|}{--} & 95.3\% & 92.1\% & 0.647 & 0.346 & 0.411 & 0.643 & \multicolumn{3}{c}{} \\
\bottomrule
\end{tabular}
\vspace{-6mm}
\end{table}

The encoder is trained on the same image-fMRI pairs available to the decoder, thus it uses no additional information from the data, yet its synthetic responses help the decoder. While related work attributes this to data augmentation or data enlargement ~\citep{mai2026synbrain,brainit}, the large improvement for concept generalization suggests the synthetic data contain additional information for \concept. 
We first test whether this concept-specific information indeed drives the
improvement, then vary the encoder's training conditions and design to locate where its
knowledge of \concept comes from, and finally ask whether standard
encoder metrics can predict the value of its synthetic data.

\paragraph{Concept-specific content of synthetic data matters.}
A simpler explanation of the gains in Sec.~\ref{sec:concept-generalization-results} is that any additional training pairs help. We test this on subject~1 with two controls that add the same number of pairs as the main condition, using encoder predictions for images without $\mathcal{C}$: \emph{non-$\mathcal{C}$ images} pairs these predictions with the images they were predicted for, adding data but no information about $\mathcal{C}$; \emph{wrong fMRI} pairs them with the $\mathcal{C}$ images instead, so the decoder sees $\mathcal{C}$ as a target but never with the response it elicits. The controls recover 2\% and 12\% of the Object Recall gap between the unseen decoder and one trained on real fMRI for $\mathcal{C}$, against 64\% for correctly paired synthetic data (Appendix~\ref{app:syn_controls}). Neither more data nor exposure to $\mathcal{C}$ images alone suffices; the gain requires the encoder's response to $\mathcal{C}$ itself.
Two further controls refine this (Appendix~\ref{app:syn_controls}). Shuffling the synthetic responses among the $\mathcal{C}$ images, which keeps supercategory-level information but breaks the pairing, removes most of the gain, so correct pairing within the concept is needed. Conversely, the pool does not have to be selected for $\mathcal{C}$: a general pool fixed in advance recovers nearly as much as targeted synthesis. What matters is whether $\mathcal{C}$ is represented in the synthesis pool, not whether the pool was selected for it. Precision also recovers together with recall, and the false-positive rate stays below that of a decoder trained on real fMRI for $\mathcal{C}$ (Appendix~\ref{app:precision}), so the gain is not explained by a tendency to generate $\mathcal{C}$ more often.

\paragraph{Comparing encoder variants.}
Table~\ref{tab:encoder-variants} defines five variants along two factors: whether \concept was available during cross-subject pre-training and during target-subject adaptation. The oracle variants observe real target-subject responses to \concept and cannot serve as synthesis sources; they are reported as upper bounds. Each of the three remaining variants is evaluated by a \emph{probe decoder}, a MindEye2 trained from scratch (to prevent cross-subject information leak from decoder pre-training) on real data plus that variant's synthetic pairs. Transfer-concept is the variant used throughout Sec.~\ref{sec:experiments}.

\paragraph{Evaluating encoders.}
\label{par:scoring-encoders}
We score each encoder in two ways. \emph{Fidelity} is the Pearson $r$ between predicted and measured fMRI. \emph{Decoded recall} measures how semantically decodable a prediction is: we reconstruct images from the encoder's predictions with a decoder, a MindEye2 trained with \concept present from the full budget plus synthetic data, and apply the concept metrics of Sec.~\ref{sec:setting}. Since the decoder has seen \concept, it can detect and recover the semantic content from the encoder's prediction. As the decoder is not a perfect judge, we intend to use this only to compare encoders (Appendix ~\ref{app:decoded-recall}).

\paragraph{Where does the encoder's knowledge of \concept come from?}
We read Table~\ref{tab:encoder-variants} first by the probe decoder, since that
is the quantity we ultimately care about. Scratch-blind, which never
sees fMRI for \concept from any subject, already recovers 41\% of
the Object Recall gap; cross-subject transfer is therefore not required,
and this baseline robustness must come from the encoder itself, which we
trace to its pretrained backbone below. At the full budget, pre-training
on other subjects without \concept (transfer-blind) adds almost
nothing: Object Recall moves from 0.218 to 0.229, and the two probe
decoders are statistically equivalent
(Appendix~\ref{app:stats-encoder}). Exposure to \concept in the
pre-training subjects (transfer-concept) is what adds: the probe decoder
improves on every metric in Table~\ref{tab:encoder-variants}, by $+0.049$ in Object Recall, and reaches 59\%
of the gap. The gain thus has two sources: the encoder's inherent
robustness to unseen concepts, and its transfer of concept exposure from
other brains (Sec.~\ref{sec:concept-generalization-results}). At the finer granularity
of single COCO categories (category generalization), every effect shrinks with the semantic gap and
the concept-specific transfer is no longer significant
(Appendix~\ref{app:encoder-category}).

\paragraph{Overall encoder metrics do not suffice to predict the value of its synthetic data.}
One might expect a more accurate encoder to yield better synthetic data. The two transfer steps above show otherwise: the step that barely helps the probe decoder, pre-training without \concept, raises Pearson $r$ by 0.060, while the step that helps on every metric, exposure to \concept, raises it by only 0.007. Pre-training makes the predicted response more accurate overall; seeing \concept, from the target subject or from others, is what puts \concept into it. Decoded recall is designed to measure the latter and does recover this ordering (scratch-oracle exceeds transfer-blind by 0.118 and matches transfer-concept), but it does not track the probe decoder either: it credits the two transfer steps about equally ($+0.089$ and $+0.119$), whereas the probe decoder credits only the second. 
Overall fidelity is thus informative but not sufficient. Across encoder designs, what tracks the decoder benefit better is the encoder's fidelity on the held-out concept itself: the CLIP score of images decoded from its predictions and within-concept identification, which needs no decoder, both order the tested encoders by their decoder benefit, whereas voxel-wise correlation does not (Appendix~\ref{app:encoder-ablation}). We present these as exploratory diagnostics rather than validated predictors.

\paragraph{The encoder generalizes where the decoder does not.}
The scratch-blind result shows that the encoder predicts useful responses to a concept it never saw in any subject's fMRI, and that a decoder trained on those predictions inherits part of that ability.
Since no fMRI for \concept enters that encoder, this robustness should come from its pretrained visual backbone: the DINOv2 representation already covers \concept, and a per-voxel readout fit on other concepts extends to it. 

Consistent with this, freezing the backbone instead of LoRA-tuning it leaves the decoder gains unchanged (Appendix~\ref{app:frozen_dino}). 
We test this explanation directly by varying the encoder design (Appendix~\ref{app:encoder-ablation}). A ridge regression from frozen DINOv2 features, with no cross-subject data and no backbone tuning, already yields useful synthetic data; replacing DINOv2 with OpenCLIP ViT-bigG, the backbone MindEye2 uses, gives similar gains; and a frozen, randomly initialized backbone gives almost none. The pretrained backbone therefore plays a major role, and the benefit is specific neither to DINOv2 nor to this encoder. It is, however, not automatic across encoder designs: synthetic data from SynBrain~\citep{mai2026synbrain}, retrained under the same holdout, gives little benefit when used as synthetic training data. We therefore find this technique useful for encoders that map a pretrained visual representation directly to voxels, but the benefit may vary depending on the encoder design.

This alone does not explain the asymmetry, since the decoders also build on pretrained components (CLIP and diffusion priors) that have seen \concept and still fail; for example, the encoder built on MindEye2's own backbone remains robust to the holdout while MindEye2 does not.
The difference is where that pretrained knowledge sits. 
For the encoder it sits at the input: \concept images already land in a representation that handles them, and only a smooth map from that representation to noisy voxels has to be learned. 
For the decoder it sits at the output: the decoder must learn which fMRI patterns to map there, and for \concept it is never shown any.
Synthetic pairs supply exactly the missing supervision, with quality bounded by the encoder. Semantic coverage is therefore an axis along which such an encoder degrades gracefully and the decoder does not, and encoder-generated data is a way to move the decoder along it.

\section{Conclusion \& Limitations}

We showed that in the concept generalization setting, in which a semantic concept is withheld from the target subject's training data, three recent fMRI-to-image decoders substantially lose their ability to reconstruct the held-out concept, even when it remains available from other subjects. Brain encoders that map a pretrained visual representation directly to voxels, trained on the same data, are far more robust to this holdout, and generating synthetic fMRI for images of the concept transfers part of that robustness to decoders: across three decoders, four subjects, and twelve concepts, it recovers 37-87\% of the gap on high-level and concept-recall metrics without any additional scanning. Our analyses indicate that this gain stems from the encoder's ability to predict responses to concepts absent from its fMRI training and from its transfer of concept exposure across subjects, rather than from added data volume.
Beyond the holdout, the same procedure improves decoding on nearly every metric at every data budget we tested, by an amount equivalent to 1.9--2.4$\times$ as many real training images. More broadly, our results suggest that decoders should be evaluated on concepts outside their training distribution, and that encoders offer a practical route to covering them.

\paragraph{Limitations.}
Our suggested solution for concept generalization has two main limitations. First, synthetic neural responses are not a perfect substitute for real ones: under an equal training budget, real fMRI for \concept still outperforms encoder-generated fMRI. Synthetic data should therefore be used when collecting real responses to \concept is impractical, not as a replacement for them. Second, synthetic pairs for images of \concept must be included in the training set. With a general synthesis pool, \concept need not be known in advance in the condition we tested, as long as the pool covers it (Sec.~\ref{sec:encoder-variants}); a pool without \concept gives no such gain, and a concept that becomes relevant only after the decoder is trained cannot be accounted for without retraining or fine-tuning.

\subsection*{Acknowledgments}
This research was supported by the MSIT (Ministry of Science, ICT), Korea, under the Global Research Support Program in the Digital Field program (RS-2024-00436680) supervised by the IITP (Institute for Information \& Communications Technology Planning \& Evaluation), and IITP grants [RS-2021-II212068, RS-2022-II220113, RS-2022-II220959, RS-2021-II211343]. This research was also supported by Microsoft Research Asia and Hyundai Motor Chung Mong-Koo Foundation.

\bibliography{iclr2027_conference}
\bibliographystyle{iclr2027_conference}

\appendix
\clearpage
\renewcommand\thesection{\Alph{section}}
\setcounter{section}{0}
{
\centering
\Large
\textbf{Concept Generalization in Visual Brain Decoding via Encoder-Generated fMRI}\\
\vspace{0.5em}Appendix \\
\vspace{1.0em}
}

\section{Additional Experimental Details}
\label{app:additional_exp_details}

\paragraph{Decoder training protocols.}
As in the original paper, MindEye2 is first pre-trained on seven other subjects and then fine-tuned to the target subject. We utilized the pre-trained MindEye2 checkpoints provided by the authors, and only fine-tuned the models ourselves. 
Brain-IT is jointly trained with all eight subjects. Brain-Diffuser is trained from scratch on the target subject as it offers no multi-subject training method. 

\paragraph{Decoder input format and the corresponding synthetic fMRI.}
Depending on the decoding model, the input fMRI voxel space may differ. While all models use single-trial GLM beta responses as the input, MindEye2 and Brain-Diffuser use the \emph{nsdgeneral} region supplied by the NSD authors which includes around 15k voxels, Brain-IT uses a larger region \citep{gifford2023algonauts}, resulting in around 40k voxels. Therefore, separate versions of encoders were trained for each fMRI voxel format, and they were used to generate appropriate fMRI inputs for each decoder. For all evaluations of the encoder outputs such as Pearson correlation, the results for the \emph{nsdgeneral} encoder (corresponding to MindEye2 and Brain-Diffuser inputs) were reported.

\paragraph{Additional details for evaluation metrics.} The low-level and high-level evaluation metrics (Inception, CLIP, EfficientNet and SwAV) follow the protocol of previous brain decoding literature: AlexNet(2), AlexNet(5), Inception and CLIP are two-way identification metrics, where the respective embeddings for the ground truth image, the reconstruction, and another reconstruction are extracted, and the percentage chance of the reconstruction being closer to the ground-truth than to another reconstruction is reported. EfficientNet and SwAV report the average correlation distance, calculated using embeddings extracted from the respective models.

\paragraph{Data counts for each supercategory.}
Table~\ref{tab:concept_counts} lists the number of training images containing \concept for each target subject (out of 9{,}000) as well as the number of test images containing \concept (out of 1{,}000). Images may contain more than one supercategory, so columns do not sum to the set size.

\paragraph{Subsampling method for limited data experiments.}
Unlike previous work, which subsamples by scan session and therefore retains an uneven number of repetitions per image, we subsample by unique image and keep all of its repetitions, giving cleaner control over the budget.
NSD shows each image three times in random order across 40 sessions, so a session-based split exposes different images a different number of times and yields training sets of varying size.

\begin{table}[t]
\centering
\caption{\textbf{Size of each concept \concept.} The 12 COCO supercategories used for supercategory generalization, each followed by the two categories (indented) used for category generalization. \textit{Person} has no underlying category. Train counts are out of 9{,}000 images per subject and test counts out of 1{,}000. Note that each image may contain several categories.}
\label{tab:concept_counts}
\small
\begin{tabular}{l rrrr r rr}
\toprule
 & \multicolumn{5}{c}{Train images containing \concept} & \multicolumn{2}{c}{Test set} \\
\cmidrule(lr){2-6} \cmidrule(lr){7-8}
Concept \concept & Subj 1 & Subj 2 & Subj 5 & Subj 7 & Mean \% & Count & \% \\
\midrule
Accessory & 1,281 & 1,237 & 1,232 & 1,211 & 13.8\% & 97 & 9.7\% \\
\quad Backpack & 393 & 411 & 413 & 404 & 4.5\% & 29 & 2.9\% \\
\quad Handbag & 465 & 478 & 473 & 446 & 5.2\% & 43 & 4.3\% \\
\addlinespace[2pt]
Animal & 1,934 & 1,929 & 1,924 & 1,943 & 21.5\% & 266 & 26.6\% \\
\quad Bird & 258 & 239 & 274 & 271 & 2.9\% & 39 & 3.9\% \\
\quad Giraffe & 222 & 208 & 215 & 219 & 2.4\% & 44 & 4.4\% \\
\addlinespace[2pt]
Appliance & 633 & 625 & 613 & 615 & 6.9\% & 66 & 6.6\% \\
\quad Oven & 233 & 229 & 216 & 214 & 2.5\% & 21 & 2.1\% \\
\quad Sink & 381 & 379 & 375 & 386 & 4.2\% & 49 & 4.9\% \\
\addlinespace[2pt]
Electronic & 839 & 867 & 852 & 783 & 9.3\% & 81 & 8.1\% \\
\quad Cell phone & 328 & 333 & 318 & 292 & 3.5\% & 31 & 3.1\% \\
\quad TV & 290 & 292 & 298 & 284 & 3.2\% & 32 & 3.2\% \\
\addlinespace[2pt]
Food & 1,194 & 1,192 & 1,211 & 1,217 & 13.4\% & 103 & 10.3\% \\
\quad Broccoli & 140 & 133 & 155 & 159 & 1.6\% & 22 & 2.2\% \\
\quad Orange & 103 & 147 & 136 & 135 & 1.4\% & 21 & 2.1\% \\
\addlinespace[2pt]
Furniture & 2,121 & 2,070 & 2,106 & 2,078 & 23.3\% & 186 & 18.6\% \\
\quad Chair & 896 & 824 & 831 & 907 & 9.6\% & 84 & 8.4\% \\
\quad Dining table & 816 & 797 & 810 & 771 & 8.9\% & 75 & 7.5\% \\
\addlinespace[2pt]
Indoor & 1,163 & 1,149 & 1,160 & 1,215 & 13.0\% & 134 & 13.4\% \\
\quad Book & 393 & 384 & 394 & 369 & 4.3\% & 40 & 4.0\% \\
\quad Clock & 368 & 382 & 392 & 408 & 4.3\% & 46 & 4.6\% \\
\addlinespace[2pt]
Kitchen & 1,508 & 1,439 & 1,487 & 1,446 & 16.3\% & 125 & 12.5\% \\
\quad Bottle & 613 & 633 & 627 & 606 & 6.9\% & 49 & 4.9\% \\
\quad Cup & 653 & 657 & 658 & 625 & 7.2\% & 56 & 5.6\% \\
\addlinespace[2pt]
Outdoor & 959 & 1,051 & 1,006 & 1,002 & 11.2\% & 101 & 10.1\% \\
\quad Bench & 402 & 466 & 427 & 413 & 4.7\% & 50 & 5.0\% \\
\quad Traffic light & 301 & 339 & 307 & 322 & 3.5\% & 35 & 3.5\% \\
\addlinespace[2pt]
Person & 4,636 & 4,567 & 4,533 & 4,488 & 50.6\% & 438 & 43.8\% \\
\addlinespace[2pt]
Sports & 1,850 & 1,744 & 1,763 & 1,773 & 19.8\% & 200 & 20.0\% \\
\quad Skis & 297 & 250 & 213 & 259 & 2.8\% & 43 & 4.3\% \\
\quad Surfboard & 309 & 319 & 321 & 303 & 3.5\% & 48 & 4.8\% \\
\addlinespace[2pt]
Vehicle & 2,073 & 2,149 & 2,111 & 2,084 & 23.4\% & 211 & 21.1\% \\
\quad Car & 876 & 927 & 919 & 884 & 10.0\% & 74 & 7.4\% \\
\quad Train & 286 & 305 & 294 & 300 & 3.3\% & 36 & 3.6\% \\
\bottomrule
\end{tabular}
\end{table}

\section{Evaluation Metrics}
\label{app:metrics}

We report the standard set of low-level and high-level reconstruction metrics used in prior work~\citep{mindeye1,mindeye2,brainit}. \emph{PixCorr} is the pixel-wise correlation between the ground-truth and the reconstructed image, and \emph{SSIM} is the grayscale structural similarity index~\citep{wang2004image}; both are low-level. \emph{Alex(2)} and \emph{Alex(5)} are two-way identification accuracies computed from the second and fifth layers of AlexNet, and \emph{Incep} and \emph{CLIP} are the corresponding accuracies from InceptionV3 pool features and the final layer of CLIP ViT-L/14, forming a sequence of increasing abstraction. \emph{Eff} and \emph{SwAV} report the average correlation distance in EfficientNet-B1~\citep{tan2019efficientnet} and SwAV-ResNet50~\citep{caron2020unsupervised} feature space, for which lower is better.

All two-way metrics ask whether a reconstruction is closer, in the relevant feature space, to its own stimulus than to an another reconstruction. They are therefore comparative, and a reconstruction that is structurally plausible but semantically empty can pass them; this is the reason we adapt the object-detection metrics of Sec.~\ref{sec:setting} for the semantic-holdout experiments.

\subsection{Object Recall and Relaxed Recall}
\label{app:recall-details}

In order to supplement standard high-level metrics that compare global embeddings and can score a reconstruction highly even when the object of interest is absent, we adapt metrics that focus on whether the target concept itself was recovered. We follow SEED~\citep{seed} and run an object detector, namely MM-Grounding-DINO~\citep{zhao2024open}, on both the ground-truth image and the reconstruction, then ask whether the detected content matches.
Formally, we index every quantity by the subject $s\in\mathcal{S}$ and the decoder variant $v\in\mathcal{V}$.
Let $\mathcal{O}$ ($|\mathcal{O}|=80$) be the set of COCO subcategories and $\mathcal{G}$ the set of $12$ COCO supercategories, which are the same supercategories as the held-out concept \concept in the main text. Each subcategory has a parent supercategory given by the map $\pi:\mathcal{O}\rightarrow\mathcal{G}$, and for a set $X\subseteq\mathcal{O}$ we write $\pi(X)=\{\pi(o):o\in X\}$.

We prompt the detector with all 80 COCO class names and, on both
ground-truth and reconstructed images, discard detections with confidence
below 0.3 or bounding-box area below 50 pixels, to filter noisy results.

\paragraph{Which model produces the reconstructions.}
Each holdout excludes an entire supercategory, so there are $12$ separately
trained models per subject and variant. Both metrics for a supercategory
$\gamma$, including the recall of every subcategory $o$ with $\pi(o)=\gamma$,
are always computed from the model trained with $\mathcal{C}=\gamma$: for
example, \emph{zebra} is scored with the animal-excluded model, \emph{bus} with
the vehicle-excluded model, and so on. This keeps every number a concept
measurement, and it is consistent with the test set restriction of
Sec.~\ref{sec:experiments}, since the test set of that model consists
precisely of images containing $\gamma$.

\paragraph{The qualifying sets.}
For a test image $i$, write $A_i\subseteq\mathcal{O}$ for object categories present in the ground-truth COCO annotation,
$D_i\subseteq\mathcal{O}$ for the detected object categories on the ground-truth
image, and $\widehat{D}_i\subseteq\mathcal{O}$ for the detected object categories on
its reconstruction.
We only score an image if the COCO annotation and the detector run on the \emph{ground-truth} image agree that the object being measured is present.
The annotation confirms the object is genuinely in the scene, and the ground-truth detection confirms the detector is able to find it without any detector-side error.
For a subcategory $o$, the qualifying set $\mathcal{Q}_o$ therefore contains the test images for which both $A_i$ and $D_i$ include $o$. For a supercategory $\gamma$, the qualifying set $\mathcal{Q}_\gamma$ contains the test images for which both $A_i$ and $D_i$ include at least one object of $\gamma$.

This condition ensures that the trivial perfect reconstruction, which is a reconstruction identical to the ground truth, always gets a perfect score for both metrics. Otherwise, even the perfect reconstruction would not get a perfect score if any detector-side error occurs.

\paragraph{The two metrics.}
\emph{Recall} is defined per subcategory: $R_o$ is the fraction of images in
$\mathcal{Q}_o$ whose reconstruction contains the exact subcategory $o$. If the
ground-truth image shows a zebra, the reconstruction must contain a zebra.
\emph{Relaxed recall} is defined directly on the concept: $\widetilde{R}_\gamma$
is the fraction of images in $\mathcal{Q}_\gamma$ whose reconstruction contains
any object of $\gamma$, so a reconstructed giraffe still counts even if the
ground truth is a zebra, as a giraffe is still an animal just like a zebra.
We define relaxed recall at the image level rather than per subcategory because
its hit condition does not depend on which subcategory is present; computing it
per subcategory would only reweight images, for instance counting an image that
contains both a zebra and a giraffe twice with the same outcome.
The relaxed variant is informative because a decoder that has never seen a
concept often produces a semantically adjacent object rather than an unrelated
one, and this near-miss behavior is invisible to exact recall. For any single
image, an exact hit is also a relaxed hit, and $\mathcal{Q}_o\subseteq\mathcal{Q}_{\pi(o)}$;
the two metrics are, however, averaged over different sets, so relaxed recall
exceeding recall is an empirical observation rather than an identity. The gap
between them indicates how much of the supercategory a decoder recovers without
pinning down the specific object.
Algorithm~\ref{alg:recall-relaxed-recall} states the full procedure.

\begin{algorithm}[h]
\caption{Object recall and relaxed recall, for a fixed subject $s$ and decoder
  variant $v$}
\label{alg:recall-relaxed-recall}
\begin{algorithmic}[1]
\Require Test images $\mathcal{I}^{\gamma}$ and reconstruction detections
  $\widehat{D}^{\gamma}_i$ for every held-out supercategory
  $\gamma\in\mathcal{G}$; annotations $A_i$ and ground-truth detections $D_i$.
\Ensure $R_o$ for every $o\in\mathcal{O}$ and $\widetilde{R}_\gamma$ for every $\gamma\in\mathcal{G}$.
\ForAll{$o\in\mathcal{O}$}
  \State $\gamma\gets\pi(o)$
    \Comment{score $o$ on the model that held out its parent}
  \State $\mathcal{Q}_o\gets\{\,i\in\mathcal{I}^{\gamma}:
    o\in A_i\ \wedge\ o\in D_i\,\}$
    \Comment{annotation and ground-truth detector agree}
  \State $R_o\gets\dfrac{1}{|\mathcal{Q}_o|}\sum_{i\in\mathcal{Q}_o}
    \mathbb{1}\!\left[o\in\widehat{D}^{\gamma}_i\right]$
    \Comment{Recall: reconstruction recovers $o$ itself}
\EndFor
\ForAll{$\gamma\in\mathcal{G}$}
  \State $\mathcal{Q}_\gamma\gets\{\,i\in\mathcal{I}^{\gamma}:
    \gamma\in\pi(A_i)\ \wedge\ \gamma\in\pi(D_i)\,\}$
    \Comment{annotation and ground-truth detector agree}
  \State $\widetilde{R}_\gamma\gets\dfrac{1}{|\mathcal{Q}_\gamma|}\sum_{i\in\mathcal{Q}_\gamma}
    \mathbb{1}\!\left[\gamma\in\pi\big(\widehat{D}^{\gamma}_i\big)\right]$
    \Comment{Relaxed Recall: reconstruction recovers $\gamma$}
\EndFor
\State \Return $\{R_o\}_{o\in\mathcal{O}}$, $\{\widetilde{R}_\gamma\}_{\gamma\in\mathcal{G}}$
\end{algorithmic}
\end{algorithm}

\paragraph{Aggregation.}
Recall is first macro-averaged over the subcategories within each
supercategory; relaxed recall is already computed per supercategory. Both are
then averaged across subjects, and finally across the $12$ supercategories with
equal weight. Weighting supercategories equally prevents frequent concepts such
as \emph{person}, which appears in nearly half of the test images, from
dominating the reported score, so that the metric reflects performance across
concepts rather than across images.

\paragraph{Category generalization.}
For the 22 categories of Table~\ref{tab:concept_counts} (two per non-person
supercategory), the procedure differs in two respects. First, each category
$o$ is scored with the model trained with $o$ itself held out, rather than
with the model that held out $\pi(o)$; an image containing several of the
22 categories is therefore scored separately by each corresponding model.
Second, both metrics use the exact-category qualifying set $Q_o$ of
Algorithm~\ref{alg:recall-relaxed-recall} (line 3), with $I^{o}$ the test images containing
$o$. Recall $R_o$ is computed as before, while relaxed recall becomes
\begin{equation}
    \widetilde R_o = \frac{1}{|Q_o|}\sum_{i\in Q_o}
    \mathbb{1}\!\left[\pi(o)\in\pi\big(\widehat D_i^{\,o}\big)\right],
\end{equation}
so a reconstruction containing a sibling $o'$ with $\pi(o')=\pi(o)$ is a
relaxed hit. Unlike in the supercategory setting, such siblings remain in
the target subject's training set, so at this granularity relaxed recall
partly credits the decoder for substituting a seen, semantically adjacent
object. Both metrics are averaged over subjects and then over the 22
categories with equal weight; since every parent contributes exactly two
categories, this is equivalent to weighting the 11 parent supercategories
equally.

\section{Additional Concept Generalization Results}
\label{app:holdout-extra}

\subsection{Complementary Test Set Results}
\label{app:complementary_results}
\begin{table}[t]
\centering
\caption{\textbf{Concept generalization for decoders, evaluated on unrelated images.} Decoders are trained with/without \concept and evaluated on images that \textbf{do not} contain \concept. Removing C's training data or adding synthetic data shifts these scores by an order of magnitude less than on images containing C; see Appendix~\ref{app:complementary_results} and Table~\ref{tab:stats_complementary}.}
\label{tab:complementary-decoder}
\small
\setlength{\tabcolsep}{2.4pt}
\begin{tabular}{llcccccccc}
\toprule
& & \multicolumn{4}{c}{Low-level} & \multicolumn{4}{c}{High-level} \\
\cmidrule(lr){3-6} \cmidrule(lr){7-10}
Decoder & Concept \concept & PixCorr $\uparrow$ & SSIM $\uparrow$ & Alex(2) $\uparrow$ & Alex(5) $\uparrow$ & Incep $\uparrow$ & CLIP $\uparrow$ & EffNet $\downarrow$ & SwAV $\downarrow$ \\
\midrule
\multirow{3}{*}[-2pt]{Brain-Diffuser} & Seen & 0.260 & 0.358 & 94.5\% & 96.6\% & 88.0\% & 92.0\% & 0.772 & 0.423 \\
\cmidrule(lr){2-10}
& Unseen & 0.255 & 0.357 & 94.1\% & 96.4\% & 87.6\% & 91.9\% & 0.773 & 0.425 \\
& Unseen + syn. & 0.260 & 0.357 & 94.6\% & 96.7\% & 87.9\% & 92.2\% & 0.770 & 0.421 \\
\midrule
\multirow{3}{*}[-2pt]{MindEye2} & Seen & 0.281 & 0.378 & 95.8\% & 98.7\% & 94.9\% & 93.7\% & 0.619 & 0.345 \\
\cmidrule(lr){2-10}
& Unseen & 0.276 & 0.379 & 95.2\% & 98.5\% & 94.6\% & 92.9\% & 0.628 & 0.350 \\
& Unseen + syn. & 0.285 & 0.380 & 95.7\% & 98.7\% & 95.0\% & 93.4\% & 0.618 & 0.344 \\
\midrule
\multirow{3}{*}[-2pt]{Brain-IT} & Seen & 0.365 & 0.502 & 98.0\% & 99.3\% & 96.8\% & 95.7\% & 0.581 & 0.336 \\
\cmidrule(lr){2-10}
& Unseen & 0.360 & 0.502 & 98.0\% & 99.2\% & 96.6\% & 95.7\% & 0.583 & 0.337 \\
& Unseen + syn. & 0.362 & 0.503 & 98.1\% & 99.2\% & 96.8\% & 95.7\% & 0.584 & 0.337 \\
\bottomrule
\end{tabular}
\vspace{-4mm}
\end{table}

Table \ref{tab:complementary-decoder} shows performance of different decoders from the concept generalization experiments in Table \ref{tab:unseen-concept-decoder}, evaluated instead on the complementary test set that do not contain \concept. In other words, while results from Table \ref{tab:unseen-concept-decoder} are from images that contain \concept, results from Table \ref{tab:complementary-decoder} are from the left-out images that do not contain \concept. Results from this table show how adding or removing data from the training set in our experiments influences semantically unrelated images that do not contain \concept.

Both manipulations have small but measurable effects on images that do not contain C (Table~\ref{tab:stats_complementary}). Removing C's images from training lowers performance on unrelated images, which is expected since the Seen model has more training data (Table ~\ref{tab:concept_counts}); the shifts are significant on all eight metrics for MindEye2, five for Brain-Diffuser, and three for Brain-IT. Adding synthetic data raises performance on unrelated images, consistent with the general benefit reported in Sec.~\ref{sec:scaling-synthetic}; this is significant on all eight metrics for MindEye2, six for Brain-Diffuser, and only PixCorr for Brain-IT. (MindEye2's SSIM moves by 0.001 in the opposite direction under the holdout, at the resolution where that metric is insensitive to all of our manipulations.)

What matters for interpreting Table 1 is the magnitude of these shifts relative to those on images containing C. On the high-level metrics, the effect of the holdout on unrelated images is 1-8\% of its effect on concept images, and the effect of synthetic data is 3-14\% for Brain-Diffuser and MindEye2 and indistinguishable from zero (not significant) for Brain-IT. The low-level metrics are less specific, as expected for a semantic manipulation: there the general effects reach around 10-30\% of the concept-specific ones. We conclude that the gap in Table 1 is predominantly a failure to generalize to C, and that the recovery from synthetic data is predominantly the encoder introducing concept-specific knowledge, with a small general improvement on top.

\subsection{Low-Level Metrics on the Target Set}
\label{app:holdout-lowlevel}

Table~\ref{tab:unseen-concept-decoder} of the main text reports only high-level and object-detection metrics on the target set, since the semantic shift is by construction a high-level one. For completeness, Table~\ref{tab:unseen-concept-decoder-lowlevel} reports the low-level metrics for the same models.

\begin{table}[t]
\centering
\caption{\textbf{Concept generalization for decoders: low-level metrics.} Same models and evaluation images as Table~\ref{tab:unseen-concept-decoder}: decoders are trained with/without \concept for the target subject and evaluated on images containing \concept. Bold marks the better of the two unseen variants, with or without synthetic data.}
\label{tab:unseen-concept-decoder-lowlevel}
\small
\setlength{\tabcolsep}{5pt}
\begin{tabular}{llcccc}
\toprule
& & \multicolumn{4}{c}{Low-level} \\
\cmidrule(lr){3-6}
Decoder & Concept \concept & PixCorr $\uparrow$ & SSIM $\uparrow$ & Alex(2) $\uparrow$ & Alex(5) $\uparrow$ \\
\midrule
\multirow{3}{*}[-2pt]{Brain-Diffuser} & Seen & 0.245 & 0.352 & 93.8\% & 96.8\% \\
\cmidrule(lr){2-6}
& Unseen & 0.221 & 0.350 & 91.7\% & 94.6\% \\
& Unseen + syn. & \textbf{0.240} & \textbf{0.351} & \textbf{93.6\%} & \textbf{96.7\%} \\
\midrule
\multirow{3}{*}[-2pt]{MindEye2} & Seen & 0.255 & 0.358 & 95.1\% & 98.6\% \\
\cmidrule(lr){2-6}
& Unseen & 0.217 & 0.353 & 92.3\% & 96.7\% \\
& Unseen + syn. & \textbf{0.244} & \textbf{0.358} & \textbf{94.0\%} & \textbf{98.1\%} \\
\midrule
\multirow{3}{*}[-2pt]{Brain-IT} & Seen & 0.337 & 0.489 & 97.6\% & 99.2\% \\
\cmidrule(lr){2-6}
& Unseen & 0.321 & \textbf{0.490} & 97.0\% & 98.7\% \\
& Unseen + syn. & \textbf{0.327} & 0.489 & \textbf{97.4\%} & \textbf{99.0\%} \\
\bottomrule
\end{tabular}
\vspace{-4mm}
\end{table}

\begin{table}[t]
\centering
\caption{\textbf{Encoder impact on decoding performance by training condition, concept generalization: low-level metrics.} Same models and evaluation images as Tables~\ref{tab:encoder-variants} and~\ref{tab:encoder-variants-category}. Each row trains a MindEye2 decoder from scratch, varying the encoder that generated its synthetic pairs. The last row of each panel uses real fMRI for \concept as an upper bound.}
\label{tab:encoder-variation-on-decoder-lowlevel}
\small
\setlength{\tabcolsep}{5pt}
\begin{tabular}{lcccc}
\toprule
& \multicolumn{4}{c}{Low-level} \\
\cmidrule(lr){2-5}
Training condition & PixCorr $\uparrow$ & SSIM $\uparrow$ & Alex(2) $\uparrow$ & Alex(5) $\uparrow$ \\
\midrule
\multicolumn{5}{l}{\emph{Supercategory generalization} (12 concepts)} \\
\midrule
\concept unseen               & 0.209 & 0.351 & 91.3\% & 96.2\% \\
+ syn.\ (scratch-blind)       & 0.232 & \textbf{0.358} & 93.0\% & 97.5\% \\
+ syn.\ (transfer-blind)      & 0.231 & 0.357 & 93.1\% & 97.5\% \\
+ syn.\ (transfer-concept)    & \textbf{0.236} & 0.358 & \textbf{93.4\%} & \textbf{97.7\%} \\
\cmidrule(l){1-5}
Real fMRI for \concept (seen) & 0.246 & 0.359 & 94.0\% & 98.3\% \\
\midrule
\multicolumn{5}{l}{\emph{Category generalization} (22 concepts)} \\
\midrule
\concept unseen               & 0.235 & 0.348 & 93.3\% & 97.5\% \\
+ syn.\ (scratch-blind)       & 0.248 & 0.350 & 94.1\% & 98.1\% \\
+ syn.\ (transfer-blind)      & 0.249 & 0.351 & \textbf{94.5\%} & 98.2\% \\
+ syn.\ (transfer-concept)    & \textbf{0.254} & \textbf{0.351} & 94.3\% & \textbf{98.3\%} \\
\cmidrule(l){1-5}
Real fMRI for \concept (seen) & 0.254 & 0.349 & 94.4\% & 98.4\% \\
\bottomrule
\end{tabular}
\end{table}

\subsection{Size-Matched Random Deletion}
\label{app:random-deletion}

Holding out a concept also shrinks the training set, by as much as half for \emph{person} (Table~\ref{tab:concept_counts}), and the comparison of Appendix~\ref{app:complementary_results} is made on a different test set. To separate the loss of \concept from the loss of training volume on the same target-concept test set, we compare concept deletion with a size-matched random deletion, using the MindEye2 probe decoder of Sec.~\ref{sec:encoder-variants} on subject~1 with \emph{animal} held out. From the full training set, we randomly remove as many images as the \emph{animal} holdout does (1{,}934), drawing only from images without animals. The resulting set matches the concept-held-out set in total real training volume, and the full set in concept-specific training volume.

Table~\ref{tab:random-deletion} shows that size-matched random deletion costs essentially nothing on animal images, while concept deletion of the same size removes almost all recall. The loss of general training volume alone therefore does not explain the deficit, consistent with the non-\concept control (Appendix~\ref{app:syn_controls}), where adding data without \concept does not help. Note that a similar experiment cannot be constructed for a very large concept like \emph{person}: it would require deleting 4{,}636 non-person images from subject~1's training set, which contains only 4{,}364.

\begin{table}[t]
\centering
\caption{\textbf{Size-matched random deletion vs.\ concept deletion.} MindEye2 probe decoder trained on real data only, subject~1, \concept{} = \emph{animal}. Random deletion removes 1{,}934 images without animals; concept deletion removes the 1{,}934 images with animals.}
\label{tab:random-deletion}
\small
\setlength{\tabcolsep}{3pt}
\begin{tabular}{lrcccccc}
\toprule
& & \multicolumn{4}{c}{Animal test images} & \multicolumn{2}{c}{Non-animal test images} \\
\cmidrule(lr){3-6}\cmidrule(lr){7-8}
Training set & \# images & Recall $\uparrow$ & Relaxed $\uparrow$ & Incep $\uparrow$ & CLIP $\uparrow$ & Incep $\uparrow$ & CLIP $\uparrow$ \\
\midrule
Full (\concept seen)                  & 9{,}000 & 0.462 & 0.748 & 88.2\% & 91.9\% & 96.5\% & 93.4\% \\
Random deletion, animals kept         & 7{,}066 & 0.456 & 0.767 & 90.3\% & 91.5\% & 95.2\% & 91.6\% \\
Concept deletion (\concept unseen)    & 7{,}066 & 0.015 & 0.054 & 72.7\% & 71.8\% & 96.1\% & 92.6\% \\
\bottomrule
\end{tabular}
\end{table}

\section{Additional Encoder-Variant Results}
\label{app:encoder-extra}

\subsection{Decoded Recall for Encoders}
\label{app:decoded-recall}

Following prior work that evaluates encoders by decoding their predictions \citep{zhang2025mindsimulator, mai2026synbrain}, we score encoders with the concept metrics of Appendix~\ref{app:recall-details}. For every test image containing \concept, we predict the target subject's response, reconstruct images from the predictions with a fixed readout, and apply Algorithm~\ref{alg:recall-relaxed-recall} unchanged. The qualifying sets depend only on annotations and ground-truth detections and are therefore identical to those used for decoders. The readout is the full-data MindEye2 for the same subject, trained with \concept present and with encoder-generated synthetic data (Sec.~\ref{sec:scaling-synthetic}), so it can read encoder outputs as well as any decoder we have. The encoder retention in Fig.~\ref{fig:retained} is the decoded recall of the transfer-concept encoder divided by that of the transfer-oracle encoder.

Because the readout's synthetic training data come from an encoder that saw \concept, the readout is expected to favor the transfer-oracle variant over the held-out ones. The retention fractions in Fig.~\ref{fig:retained} are therefore conservative, and the decoded columns of Table~\ref{tab:encoder-variants} should be read for ordering rather than for the size of differences between the scratch and transfer families.

\paragraph{Scope of the comparison and a real-data-only readout.}
The readout is intentionally trained with \concept: when an encoder does produce a response that carries the concept, the readout has to convert it into an image containing \concept, and a readout that never saw \concept would add its own concept-generalization limits to the measurement. The retention ratios of Fig.~\ref{fig:retained} nevertheless compare two fixed pipelines, which differ in their input, noise and readout, and do not isolate an intrinsic advantage of the encoding direction; we use them as motivation for the method, whose utility is tested by training decoders (Sec.~\ref{sec:experiments}).

To check that the pattern does not depend on the readout's synthetic training, we repeat the analysis on subject~1 with \emph{animal} held out, using a MindEye2 readout trained on real data only (Table~\ref{tab:readout-real}). Each encoder's retention is relative to its own oracle, decoded by the same readout. The pattern is unchanged: the encoder retains most of its recall under the holdout, while the decoder retains almost none. Retention is graded across encoder variants: it is highest with cross-subject exposure to \concept, and still about half for scratch-blind, which never saw fMRI for \concept from any subject. The oracles exceed real test fMRI, possibly because encoder predictions are free of trial noise.

\begin{table}[t]
\centering
\caption{\textbf{Encoder retention with a real-data-only readout.} Subject~1, \concept{} = \emph{animal}. Encoder-predicted test fMRI is decoded by a MindEye2 readout trained with \concept present on real data only. \emph{Retained} is the ratio to the oracle of the same family (Recall / Relaxed Recall). The last row is the probe decoder of Sec.~\ref{sec:encoder-variants} on real test fMRI, relative to its own \concept-seen model (Table~\ref{tab:encoder-ablation}).}
\label{tab:readout-real}
\small
\setlength{\tabcolsep}{3pt}
\begin{tabular}{llccccc}
\toprule
Encoder & \concept in encoder training & Recall $\uparrow$ & Relaxed $\uparrow$ & Incep $\uparrow$ & CLIP $\uparrow$ & Retained \\
\midrule
Real test fMRI (no encoder) & --                          & 0.484 & 0.771 & 96.1\% & 93.4\% & -- \\
\midrule
Transfer-oracle             & seen                        & 0.559 & 0.810 & 97.0\% & 94.0\% & -- \\
Transfer-concept            & held out (target subject)   & 0.471 & 0.713 & 96.4\% & 93.4\% & 84\% / 88\% \\
Transfer-blind              & held out (all subjects)     & 0.358 & 0.663 & 96.0\% & 92.9\% & 64\% / 82\% \\
\midrule
Scratch-oracle              & seen                        & 0.535 & 0.798 & 95.8\% & 93.5\% & -- \\
Scratch-blind               & held out                    & 0.262 & 0.593 & 93.9\% & 91.7\% & 49\% / 74\% \\
\midrule
Decoder (probe)             & held out                    & 0.015 & 0.054 & 72.7\% & 71.8\% & 3\% / 7\% \\
\bottomrule
\end{tabular}
\end{table}

\begin{table}[t]
\centering
\caption{\textbf{Synthetic-data controls and frozen-backbone encoders.}
MindEye2 probe decoder trained from scratch on subject~1, supercategory
generalization (12 concepts). Both controls use identical synthetic
fMRI, predicted for images not containing \concept; they differ
only in whether it is paired with those images (non-\concept
images) or with the \concept images (wrong fMRI). \emph{Frozen}
encoders keep DINOv2 fixed instead of LoRA-tuning it. The last row uses
real fMRI for \concept as an upper bound.}
\label{tab:controls_frozen}
\resizebox{\linewidth}{!}{%
\begin{tabular}{lcccccccccc}
\toprule
& \multicolumn{4}{c}{Low-level} & \multicolumn{4}{c}{High-level} & \multicolumn{2}{c}{Object Detection} \\
\cmidrule(lr){2-5} \cmidrule(lr){6-9} \cmidrule(lr){10-11}
Training condition & PixCorr $\uparrow$ & SSIM $\uparrow$ & Alex(2) $\uparrow$ & Alex(5) $\uparrow$ & Incep $\uparrow$ & CLIP $\uparrow$ & EffNet $\downarrow$ & SwAV $\downarrow$ & Recall $\uparrow$ & Relaxed $\uparrow$ \\
\midrule
\concept unseen                             & 0.247 & 0.360 & 94.0\% & 97.1\% & 88.4\% & 82.9\% & 0.770 & 0.411 & 0.080 & 0.219 \\
\midrule
\quad + syn.\ (non-\concept images)         & 0.262 & 0.365 & 94.2\% & 97.6\% & 88.1\% & 83.7\% & 0.766 & 0.407 & 0.088 & 0.242 \\
\quad + syn.\ (wrong fMRI)                       & 0.238 & 0.356 & 92.5\% & 96.4\% & 87.7\% & 83.1\% & 0.769 & 0.415 & 0.122 & 0.298 \\
\midrule
\quad + syn.\ (scratch-blind)                    & 0.269 & 0.367 & 95.1\% & 98.1\% & 91.6\% & 87.2\% & 0.718 & 0.383 & 0.228 & 0.435 \\
\quad + syn.\ (transfer-blind)                   & 0.272 & 0.364 & 95.2\% & 98.2\% & 92.1\% & 87.8\% & 0.718 & 0.381 & 0.222 & 0.432 \\
\quad + syn.\ (transfer-blind, frozen)           & 0.272 & 0.361 & 95.2\% & 98.3\% & 92.5\% & 87.5\% & 0.717 & 0.379 & 0.239 & 0.447 \\
\quad + syn.\ (transfer-concept)                 & 0.276 & 0.365 & 95.5\% & 98.5\% & 93.2\% & 89.0\% & 0.698 & 0.371 & 0.301 & 0.519 \\
\quad + syn.\ (transfer-concept, frozen)         & 0.275 & 0.364 & 95.4\% & 98.5\% & 93.1\% & 89.2\% & 0.702 & 0.374 & 0.287 & 0.515 \\
\midrule
Real fMRI for \concept (seen)               & 0.282 & 0.364 & 95.9\% & 98.9\% & 95.6\% & 92.7\% & 0.638 & 0.338 & 0.423 & 0.644 \\
\bottomrule
\end{tabular}}
\end{table}

\subsection{Category generalization}
\label{app:encoder-category}

To probe a smaller semantic gap than a supercategory, we repeat the analysis of Sec.~\ref{sec:encoder-variants} with individual COCO categories as \concept, taking the two most frequent categories of each supercategory among the 1{,}000 test set images; \emph{person} is excluded as it has no subcategories, giving 22 categories (Table~\ref{tab:concept_counts}). Table~\ref{tab:encoder-variants-category} reports the same quantities as Table~\ref{tab:encoder-variants}.

\begin{table}[t]
\centering
\caption{\textbf{Encoder variants and their value as synthetic-data sources, category generalization (22 concepts).} Same layout as Table~\ref{tab:encoder-variants}. Bold marks the best of the three synthesis encoders in each probe-decoder column.}
\label{tab:encoder-variants-category}
\footnotesize
\setlength{\tabcolsep}{2.5pt}
\begin{tabular}{lcc|cccccc|ccc}
\toprule
& \multicolumn{2}{c|}{\concept{} incl.\ for} & \multicolumn{6}{c|}{Probe decoder trained with synthetic data} & \multicolumn{3}{c}{Encoder} \\
\cmidrule(lr){2-3}\cmidrule(lr){4-9}\cmidrule(lr){10-12}
Encoder variant & Other & Target & Incep $\uparrow$ & CLIP $\uparrow$ & EffNet $\downarrow$ & SwAV $\downarrow$ & Rec.\ $\uparrow$ & Rel.\ $\uparrow$ & $r$ $\uparrow$ & D.Rec.\ $\uparrow$ & D.Rel.\ $\uparrow$ \\
\midrule
Scratch-oracle   & --        & \ding{51} & \multicolumn{6}{c|}{\multirow{2}{*}{\textit{n/a: trained on real fMRI for \concept}}} & 0.275 & 0.551 & 0.693 \\
Transfer-oracle  & \ding{51} & \ding{51} & \multicolumn{6}{c|}{}                                                                  & 0.322 & 0.623 & 0.738 \\
\midrule
Scratch-blind    & --        & \ding{55} & 94.5\% & 91.9\% & 0.685 & 0.363 & 0.338 & \textbf{0.582} & 0.268 & 0.436 & 0.644 \\
Transfer-blind   & \ding{55} & \ding{55} & 94.7\% & \textbf{92.4\%} & \textbf{0.675} & 0.360 & 0.351 & 0.580 & 0.318 & 0.513 & 0.692 \\
Transfer-concept & \ding{51} & \ding{55} & \textbf{94.8\%} & 92.1\% & 0.676 & \textbf{0.358} & \textbf{0.365} & 0.578 & 0.320 & 0.551 & 0.697 \\
\midrule
No synthetic     & & & 92.4\% & 90.2\% & 0.729 & 0.383 & 0.179 & 0.505 & & & \\
Real fMRI (seen) & & & 96.0\% & 93.2\% & 0.636 & 0.339 & 0.469 & 0.650 & & & \\
\bottomrule
\end{tabular}
\end{table}

Every effect shrinks with the semantic gap. Removing \concept from all subjects costs the encoder 0.004 in $r$ (1\%) and 18\% of decoded Object Recall, against 0.014 and 34\% at supercategory granularity. The reversal between fidelity and decoded recall for the pre-training contrast (scratch-oracle vs.\ transfer-blind) shrinks to 0.038 in Object Recall and is not robust across tests (Appendix~\ref{app:stats-encoder}). For the probe decoder, concept-specific transfer is significant on no metric at this granularity, whereas it is significant on every metric at supercategory granularity; the dependence on granularity is itself significant for Inception, CLIP and EffNet. 
The within-cell rank correlation between encoder $r$ and probe-decoder Object Recall vanishes ($\rho = 0.10$, n.s.). 
Scratch-blind and transfer-blind still recover 55\% and 59\% of the Object Recall gap. 
Note that at this granularity sibling categories of \concept remain in training, so relaxed recall partly credits the decoder for substituting a seen, semantically adjacent object (Appendix~\ref{app:metrics}).

\subsection{Synthetic-data controls}
\label{app:syn_controls}

Sec.~\ref{sec:encoder-variants} attributes the benefit of synthetic data to the
encoder's ability to predict responses to \concept. A simpler
explanation is that any additional training pairs help, as in
conventional data augmentation. We test this with two controls that
add the same number of synthetic pairs as the main condition and use
identical synthetic responses, differing only in which images those
responses are paired with.

\paragraph{Non-\concept images.} The synthetic responses are
predicted by the transfer-concept encoder for images that do not
contain \concept, and are paired with those same images. The
pairs are therefore correctly matched but carry no information about
\concept, which isolates the effect of training-set size.

\paragraph{Wrong fMRI.} The same responses are instead paired with the
\concept images that the main condition uses. The decoder thus
sees \concept as a reconstruction target, but never with the
response that \concept elicits. Because the fMRI is identical to
the non-\concept control and comes from the same encoder, any
failure of this control cannot be attributed to the synthetic
responses being off-distribution.

\paragraph{Results.} Table~\ref{tab:controls_frozen} reports both
controls for subject~1. Neither closes the gap: both fall well short of
correctly paired synthetic data on Object Recall ($-0.21$ and $-0.18$)
and Relaxed Recall ($-0.28$ and $-0.22$; all $p < 0.001$). The
non-\concept control is statistically equivalent to the unseen
baseline on every high-level and object-detection metric (Object Recall
0.080 to 0.088), yet significantly improves the low-level metrics
(PixCorr $+0.015$, SSIM $+0.005$, Alex(5) $+0.4$ points), consistent
with the general benefit of synthetic data in Sec.~\ref{sec:scaling-synthetic}.
Additional data therefore improves decoding in general but not on
\concept. Wrong fMRI instead lowers Alex(2), Alex(5) and SSIM
significantly, as expected from mismatched pairs acting as label noise,
and modestly raises Object Recall (0.080 to 0.122) and Relaxed Recall
(0.219 to 0.298). This is consistent with the decoder's output prior
drifting toward \concept once \concept images appear as
targets, but it amounts to only 19\% [12, 26] and 27\% [19, 35] of the
gain from correct pairing. The benefit of synthetic data therefore
requires both that the images depict \concept and that they are
paired with the encoder's prediction of the response to them. Neither
additional data volume nor exposure to \concept images alone is
sufficient.

\paragraph{Within-concept shuffling and a fixed general pool.}
The wrong-fMRI control pairs \concept images with responses to images without \concept, so it cannot tell whether the remaining advantage of correct pairing comes from coarse supercategory information or from image-specific information. Both controls above also leave open whether the synthesis pool must be selected for \concept. We therefore add two controls on subject~1 with \emph{animal} held out (Table~\ref{tab:general-pool}).

\emph{Shuffled within \concept} pairs each \concept image with the synthetic response of a different \concept image, which keeps supercategory-level information but breaks the pairing. It raises only Relaxed Recall (0.054 to 0.209), while the other metrics stay similar to or fall below the unseen baseline. Supercategory-level supervision alone is therefore not enough: correct pairing within the concept is needed.

\emph{Fixed general pool} replaces the \concept images with a pool that is fixed in advance and not selected for the test concept: the entire training set of subject~2 (9{,}000 images, of which 1{,}929 contain animals, against 1{,}934 held out from subject~1). It recovers nearly as much of the gap as targeted synthesis. This pool is larger than the main setting (9{,}000 against 1{,}934 synthetic images), but added volume alone does not help on \concept, as the non-\concept control shows. Together, the two results indicate that what matters is whether \concept is represented in the synthesis pool, not whether the pool was selected for it: a broad pool can cover a held-out concept without knowing it in advance, as long as the concept appears in it. Since a held-out supercategory removes several COCO categories jointly (e.g., zebra, dog and bird for \emph{animal}) and Object Recall is scored per category, one decoder and one untargeted pool recover several held-out categories at once. Targeting the pool remains useful when the concept is known, as it reaches a similar performance with far fewer synthetic pairs.

\begin{table}[t]
\centering
\caption{\textbf{Within-concept shuffling and a fixed general pool.} MindEye2 probe decoder, subject~1, \concept{} = \emph{animal}, synthetic data from the transfer-concept encoder. Gap rec.\ is the recovered fraction of the gap, $(\text{syn} - \text{unseen})/(\text{seen} - \text{unseen})$, for Recall / Relaxed Recall.}
\label{tab:general-pool}
\small
\setlength{\tabcolsep}{5pt}
\begin{tabular}{lccccc}
\toprule
Synthetic data & Incep $\uparrow$ & CLIP $\uparrow$ & Recall $\uparrow$ & Relaxed $\uparrow$ & Gap rec. \\
\midrule
None (\concept unseen)            & 72.7\% & 71.8\% & 0.015 & 0.054 & -- \\
\midrule
\concept images (main setting)    & 83.4\% & 86.4\% & 0.270 & 0.543 & 57\% / 70\% \\
Fixed general pool                & 83.3\% & 88.1\% & 0.234 & 0.543 & 49\% / 70\% \\
Shuffled within \concept          & 66.3\% & 70.4\% & 0.029 & 0.209 & 3\% / 22\% \\
\midrule
Real fMRI for \concept (seen)     & 88.2\% & 91.9\% & 0.462 & 0.748 & -- \\
\bottomrule
\end{tabular}
\end{table}

\subsection{Frozen encoder backbone}
\label{app:frozen_dino}

The Universal Encoder adapts its DINOv2 backbone with LoRA during
pre-training. If the encoder's robustness to \concept came from this
fMRI-driven adaptation rather than from the pre-trained representation,
freezing the backbone should reduce it. We retrain the transfer-blind
and transfer-concept encoders with DINOv2 frozen, so that fMRI training
updates only the remaining encoder parameters, and train the probe
decoder on their synthetic pairs (Table~\ref{tab:controls_frozen}).

Freezing the backbone has no detectable effect on the decoder gains
from either encoder. For transfer-blind, Object Recall is 0.222 with
LoRA and 0.239 frozen (Relaxed Recall 0.432 and 0.447); for
transfer-concept, it is 0.301 and 0.287 (0.519 and 0.515). Both frozen
encoders are statistically equivalent to their LoRA counterparts on all
four high-level metrics and both object-detection metrics, with 90\%
confidence intervals for the Object Recall difference within
$\pm 0.033$ (Appendix~\ref{app:stats_frozen}). The low-level metrics
are mostly inconclusive rather than different; the only significant
difference is a small SSIM decrease for transfer-blind ($-0.003$).
Since transfer-blind never observes \concept from any subject, its
ability to produce useful responses to \concept must come from the
pre-trained representation itself. The concept-specific advantage of
transfer-concept over transfer-blind also persists with a frozen
backbone (Object Recall $+0.049$ [0.028, 0.070], Relaxed Recall
$+0.068$ [0.039, 0.097], both $p < 0.001$), so concept-specific
transfer through other subjects does not require adapting DINOv2.
Whether freezing modestly reduces this advantage is unresolved (the
difference in advantage is $0.030$ [0.000, 0.059] in Object Recall).
Equivalence holds at the margin used throughout the paper but not at a
stricter margin of half the concept-specific advantage, so we cannot
exclude differences of that size. These results are for a single
subject and do not account for variation across training runs.
Appendix~\ref{app:encoder-ablation} tests the role of the backbone more directly.

\subsection{Encoder designs as synthetic-data sources}
\label{app:encoder-ablation}

All experiments so far use the Universal Encoder (UE). To test whether the conclusions depend on this encoder or on its DINOv2 backbone, we keep the MindEye2 probe decoder of Sec.~\ref{sec:encoder-variants} and vary the encoder that generates its synthetic data, on subject~1 with \emph{animal} held out. Each condition is trained once, for a single subject and concept, so the results in this subsection are reported without significance tests.

\paragraph{Encoders.}
Besides the three UE variants of Table~\ref{tab:encoder-variants}, we test four encoders: a ridge regression from frozen DINOv2 features to voxels, fit only on the target subject's training data; a UE with a frozen, randomly initialized ViT backbone; a UE whose backbone is replaced with OpenCLIP ViT-bigG, the backbone MindEye2 uses; and SynBrain~\citep{mai2026synbrain}, which we reproduce with the authors' official code and recipe, verify against the reported performance and the released checkpoint, and then retrain under the same holdout.

\paragraph{Results.}
Table~\ref{tab:encoder-ablation} reports the probe decoder under each encoder. The ridge encoder, whose representation was never exposed to fMRI, already produces useful synthetic data, at roughly half the benefit of the UE trained on the same subject's data (scratch-blind). The benefit therefore requires neither a sophisticated encoder nor cross-subject pre-training. Replacing DINOv2 with OpenCLIP gives similar gains, while a randomly initialized backbone gives almost none: the pretrained backbone plays a major role, and the benefit is not specific to DINOv2. The OpenCLIP result also sharpens the asymmetry of Sec.~\ref{sec:encoder-variants}: built on the same backbone as MindEye2, the encoder remains robust to the holdout while MindEye2 does not, consistent with the asymmetry depending on where the pretrained representation sits (encoder input vs.\ decoder output) rather than on which representation is used.

\paragraph{SynBrain under the holdout.}
Not every encoder shares this property: SynBrain's synthetic data gives little recovery. Decoding each encoder's predicted test fMRI with the fixed readout of Sec.~\ref{sec:encoder-variants} (Table~\ref{tab:synbrain}) shows that, under the holdout, SynBrain's predictions for the held-out concept degrade while those for seen concepts stay intact. Its predictions also carry less within-concept information than the UE's, both when the UE is matched on training data and when it is built on the same OpenCLIP backbone. One possible explanation is SynBrain's VAE voxel decoder, which is trained only on seen-concept fMRI, although establishing this would require a matched architectural ablation. SynBrain was designed for in-distribution fidelity and variability; our finding is specific to concept holdout and shows that useful concept generalization is not automatic across encoder architectures.

\paragraph{Encoder-side diagnostics.}
Table~\ref{tab:encoder-diagnostics} compares encoder-side metrics with the decoder gap recovered. Besides voxel-wise correlation and decoded recall (Sec.~\ref{sec:encoder-variants}), we test two metrics that do not require training a decoder with synthetic data. \emph{Decoded CLIP} is computed like decoded recall, but reports the CLIP metric of the reconstructed images. \emph{Within-concept identification} needs no decoder at all: it measures how often an encoder's prediction is matched to the correct image among the held-out concept images from training, after subtracting the mean response over all images.

Overall voxel-wise correlation does not consistently capture differences in synthetic-data utility, as encoders with similar correlation can differ widely in their impact on the decoder. Decoded CLIP and within-concept identification order the tested encoders by their decoder benefit, and decoded recall agrees except for the two weakest encoders. Consistent with Appendix~\ref{app:stats-encoder}, overall fidelity is informative but not sufficient, especially across encoder families; what separates the encoders is their fidelity on a concept held out from their training. In practice, both diagnostics can be computed by holding out a proxy concept for which the subject has measured responses. We present them as exploratory diagnostics rather than validated predictors, and a general-purpose metric for decoding and other downstream tasks remains future work.

\begin{table}[t]
\centering
\caption{\textbf{Encoder designs as synthetic-data sources.} MindEye2 probe decoder trained with synthetic data from each encoder, subject~1, \concept{} = \emph{animal}. Gap rec.\ is $(\text{syn} - \text{unseen})/(\text{seen} - \text{unseen})$ for Recall / Relaxed Recall. The first two rows are the probe decoder trained with real fMRI for \concept and without synthetic data. OpenCLIP denotes the OpenCLIP ViT-bigG backbone.}
\label{tab:encoder-ablation}
\footnotesize
\setlength{\tabcolsep}{3pt}
\begin{tabular}{lccccccc}
\toprule
Encoder for synthesis & Incep $\uparrow$ & CLIP $\uparrow$ & EffNet $\downarrow$ & SwAV $\downarrow$ & Recall $\uparrow$ & Relaxed $\uparrow$ & Gap rec. \\
\midrule
\concept seen (upper bound)              & 88.2\% & 91.9\% & 0.666 & 0.365 & 0.462 & 0.748 & -- \\
\concept unseen (no synthetic)           & 72.7\% & 71.8\% & 0.898 & 0.511 & 0.015 & 0.054 & -- \\
\midrule
UE, frozen random-init ViT               & 73.0\% & 76.8\% & 0.886 & 0.506 & 0.028 & 0.147 & 3\% / 13\% \\
Ridge, frozen DINOv2                     & 75.5\% & 79.7\% & 0.846 & 0.474 & 0.094 & 0.275 & 18\% / 32\% \\
SynBrain                                 & 72.0\% & 73.0\% & 0.887 & 0.505 & 0.026 & 0.093 & 2\% / 6\% \\
\midrule
UE scratch-blind                         & 79.7\% & 80.7\% & 0.810 & 0.453 & 0.167 & 0.434 & 34\% / 55\% \\
UE transfer-blind                        & 79.5\% & 85.1\% & 0.786 & 0.433 & 0.212 & 0.504 & 44\% / 65\% \\
UE transfer-concept (DINOv2)             & 83.4\% & 86.4\% & 0.767 & 0.422 & 0.270 & 0.543 & 57\% / 70\% \\
UE transfer-concept (OpenCLIP)           & 81.1\% & 84.2\% & 0.782 & 0.435 & 0.253 & 0.484 & 53\% / 62\% \\
\bottomrule
\end{tabular}
\end{table}

\begin{table}[t]
\centering
\caption{\textbf{Decoding encoder-predicted test fMRI under the holdout.} Subject~1, \concept{} = \emph{animal}; Incep / CLIP of images reconstructed from each encoder's predictions by the fixed MindEye2 readout of Sec.~\ref{sec:encoder-variants}. Within-concept identification is defined in Appendix~\ref{app:encoder-ablation}. The OpenCLIP UE uses OpenCLIP ViT-bigG, the same backbone as SynBrain.}
\label{tab:synbrain}
\small
\setlength{\tabcolsep}{3pt}
\begin{tabular}{llccc}
\toprule
Encoder & Animals in training? & \makecell{Animal test\\($n{=}266$)} & \makecell{Non-animal test\\($n{=}734$)} & \makecell{Within-concept\\identification} \\
\midrule
SynBrain, released checkpoint          & Yes              & 94.3\% / 93.7\% & 98.1\% / 95.6\% & -- \\
SynBrain, our retrain                  & No               & 82.2\% / 80.0\% & 98.0\% / 96.2\% & 20\% \\
\midrule
UE scratch-blind (subject~1 only)      & No               & 87.3\% / 88.6\% & 98.6\% / 96.3\% & 30\% \\
UE transfer-blind                      & No (any subject) & 93.7\% / 94.6\% & 99.1\% / 97.3\% & 58\% \\
UE transfer-blind (OpenCLIP)           & No (any subject) & 85.1\% / 84.1\% & 98.8\% / 96.4\% & 38\% \\
\bottomrule
\end{tabular}
\end{table}

\begin{table}[t]
\centering
\caption{\textbf{Encoder-side metrics vs.\ value of synthetic data.} Subject~1, \concept{} = \emph{animal}. The first four columns evaluate the encoder-predicted fMRI (decoded metrics on the animal test images); the last column is the decoder gap recovered when the synthetic data is used for training (Table~\ref{tab:encoder-ablation}).}
\label{tab:encoder-diagnostics}
\small
\setlength{\tabcolsep}{4pt}
\begin{tabular}{lccccc}
\toprule
Encoder & \makecell{Voxel-wise\\$r$ $\uparrow$} & \makecell{Decoded\\CLIP $\uparrow$} & \makecell{Decoded\\Recall $\uparrow$} & \makecell{Within-concept\\identification $\uparrow$} & \makecell{Decoder gap rec.\\(Recall / Relaxed)} \\
\midrule
SynBrain               & 0.334 & 80.0\% & 0.14 & 20\% & 2\% / 6\% \\
Ridge, frozen DINOv2   & 0.362 & 82.9\% & 0.11 & 28\% & 18\% / 32\% \\
UE scratch-blind       & 0.342 & 88.6\% & 0.26 & 30\% & 34\% / 55\% \\
UE transfer-blind      & 0.398 & 94.6\% & 0.42 & 58\% & 44\% / 65\% \\
UE transfer-concept    & 0.400 & 96.1\% & 0.58 & 62\% & 57\% / 70\% \\
\bottomrule
\end{tabular}
\end{table}

\subsection{Precision and false-positive rate}
\label{app:precision}

Object Recall and Relaxed Recall are scored on images that contain \concept, and do not penalize a decoder for generating \concept where it is absent. To check that the gains are not explained by a tendency to generate \concept more often, we additionally report precision and false-positive rate (FPR) for the probe decoders of Table~\ref{tab:encoder-variants} (Table~\ref{tab:precision}). Precision is the fraction of the reconstructions in which a category is detected whose ground-truth image also contains it, and FPR is the fraction of the test images without the category whose reconstruction nevertheless contains it. All metrics are computed over the full test set, including images without \concept, with the same models and averaging as Table~\ref{tab:encoder-variants}; the relaxed versions are at the supercategory level. Precision is undefined when no reconstruction contains the category (20\% of subject-category pairs for the model without synthetic data, 3--7\% for the others), so it is averaged over the 246 of 320 pairs defined for all five variants.

Precision recovers together with recall. The FPR does rise relative to the model without synthetic data, which rarely generates \concept at all, but it stays below that of the decoder trained on real fMRI for \concept. Decoders trained with synthetic data therefore do not produce \concept on images without it more often than a decoder that has seen real data for \concept.

\begin{table}[t]
\centering
\caption{\textbf{Precision and false-positive rate on held-out concepts.} Probe decoders of Table~\ref{tab:encoder-variants}, same models and averaging. Relaxed versions are computed at the supercategory level.}
\label{tab:precision}
\small
\setlength{\tabcolsep}{4pt}
\begin{tabular}{lcccccc}
\toprule
& \multicolumn{2}{c}{Recall $\uparrow$} & \multicolumn{2}{c}{Precision $\uparrow$} & \multicolumn{2}{c}{FPR $\downarrow$} \\
\cmidrule(lr){2-3}\cmidrule(lr){4-5}\cmidrule(lr){6-7}
Training condition & Exact & Relaxed & Exact & Relaxed & Exact & Relaxed \\
\midrule
Real fMRI for \concept        & 0.411 & 0.643 & 0.461 & 0.686 & 2.01\% & 4.69\% \\
No synthetic                  & 0.087 & 0.228 & 0.273 & 0.573 & 1.23\% & 3.13\% \\
\midrule
+ syn.\ (transfer-concept)    & 0.279 & 0.486 & 0.416 & 0.680 & 1.58\% & 3.82\% \\
+ syn.\ (transfer-blind)      & 0.229 & 0.441 & 0.397 & 0.671 & 1.50\% & 3.60\% \\
+ syn.\ (scratch-blind)       & 0.218 & 0.427 & 0.393 & 0.658 & 1.56\% & 3.74\% \\
\bottomrule
\end{tabular}
\end{table}

\section{Additional Data-scaling Results}

\subsection{Limited and full data results written out}

Table~\ref{tab:full_limited_results} lists part of the exact results for the experiment in Sec.~\ref{sec:scaling-synthetic}, with values identical to Fig.~\ref{fig:various_data_scale}.

\begin{table}[t]
\centering
\caption{\textbf{Effect of synthetic data across data scales}. Numbers correspond to values from Figures ~\ref{fig:various_data_scale} and~\ref{fig:various_data_scale_appendix}. The better result within each decoder pair in \textbf{bold}.}

\label{tab:full_limited_results}
\small
\setlength{\tabcolsep}{1pt}
\begin{tabular}{llcccccccc}
\toprule
& & \multicolumn{4}{c}{Low-level} & \multicolumn{4}{c}{High-level} \\
\cmidrule(lr){3-6}\cmidrule(lr){7-10}
Decoder & Syn. & PixCorr $\uparrow$ & SSIM $\uparrow$ & Alex(2) $\uparrow$ & Alex(5) $\uparrow$ & Incep $\uparrow$ & CLIP $\uparrow$ & EffNet $\downarrow$ & SwAV $\downarrow$ \\
\midrule
\multicolumn{10}{l}{\emph{Full data} ($9{,}000$ images)} \\
\midrule
\multirow{2}{*}{Brain-Diffuser} & \ding{55} & 0.258 & \textbf{0.357} & 94.4\% & 96.6\% & 88.2\% & 91.9\% & 0.770 & 0.420 \\
& \ding{51} & \textbf{0.273} & 0.353 & \textbf{95.5\%} & \textbf{97.4\%} & \textbf{90.1\%} & \textbf{93.3\%} & \textbf{0.749} & \textbf{0.401} \\
\midrule
\multirow{2}{*}{MindEye2} & \ding{55} & 0.278 & 0.376 & 95.6\% & 98.6\% & 94.9\% & 93.4\% & 0.621 & 0.344 \\
& \ding{51} & \textbf{0.312} & \textbf{0.389} & \textbf{96.9\%} & \textbf{99.1\%} & \textbf{96.1\%} & \textbf{94.1\%} & \textbf{0.604} & \textbf{0.330} \\
\midrule
\multirow{2}{*}{Brain-IT} & \ding{55} & 0.362 & \textbf{0.501} & 98.0\% & 99.3\% & 96.8\% & 95.6\% & 0.584 & 0.336 \\
& \ding{51} & \textbf{0.371} & 0.494 & \textbf{98.7\%} & \textbf{99.5\%} & \textbf{97.2\%} & \textbf{95.9\%} & \textbf{0.566} & \textbf{0.327} \\
\midrule
\multicolumn{10}{l}{\emph{Limited data} ($225$ images)} \\
\midrule
\multirow{2}{*}{Brain-Diffuser} & \ding{55} & 0.128 & 0.322 & 77.7\% & 82.4\% & 71.8\% & 77.4\% & 0.899 & 0.533 \\
& \ding{51} & \textbf{0.164} & \textbf{0.335} & \textbf{83.5\%} & \textbf{87.6\%} & \textbf{76.2\%} & \textbf{81.6\%} & \textbf{0.874} & \textbf{0.507} \\
\midrule
\multirow{2}{*}{MindEye2} & \ding{55} & 0.148 & 0.360 & 82.4\% & 89.2\% & 79.5\% & 79.9\% & 0.822 & 0.476 \\
& \ding{51} & \textbf{0.184} & \textbf{0.375} & \textbf{88.3\%} & \textbf{93.4\%} & \textbf{84.0\%} & \textbf{83.1\%} & \textbf{0.778} & \textbf{0.442} \\
\midrule
\multirow{2}{*}{Brain-IT} & \ding{55} & 0.149 & 0.479 & 79.2\% & 86.2\% & 80.7\% & 79.7\% & 0.822 & 0.509  \\
& \ding{51} & \textbf{0.174} & \textbf{0.485} & \textbf{81.1\%} & \textbf{87.8\%} & \textbf{82.3\%} & \textbf{81.2\%} & \textbf{0.806} & \textbf{0.495} \\

\bottomrule
\end{tabular}
\end{table}

\subsection{Encoder pre-training across data budgets}
\label{app:encoder-pretrain-scales}

\begin{figure}[t]
  \centering
  \includegraphics[width=\linewidth]{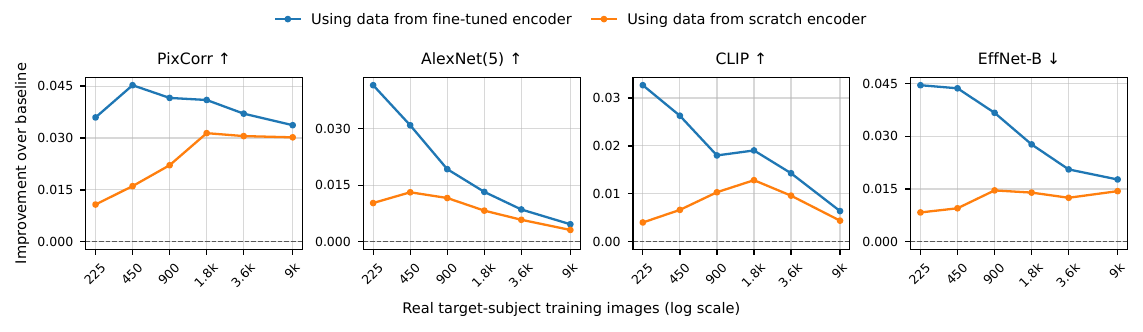}\\[0.5em]
  \includegraphics[width=\linewidth]{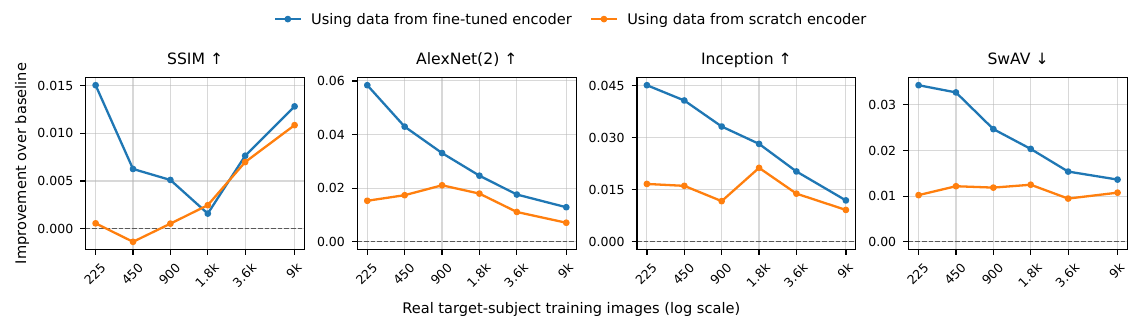}
  \caption{\textbf{Encoder pre-training matters most when target-subject data is scarce.}
  Improvement of synthetic data for MindEye2 over its real-data-only baseline, on all eight metrics, when the synthetic data come from a cross-subject pre-trained encoder (blue) versus an encoder trained from scratch on the target subject (orange). Both help at every budget, but the pre-trained encoder's advantage is largest at small budgets and shrinks as target-subject data grows (Appendix ~\ref{app:stats_encoder_budget}). SSIM is the exception, with small gains and no consistent ordering between the two encoders.}
  \label{fig:encoder-pretrain-scales}
\end{figure}

 Figure~\ref{fig:encoder-pretrain-scales} investigates the effect of encoder pre-training across budgets, using the full MindEye2 of Sec.~\ref{sec:scaling-synthetic} with synthetic data from either a cross-subject pre-trained and fine-tuned encoder or one trained from scratch on the target subject. A scratch encoder delivers roughly a quarter of the pre-trained encoder's improvement at 225 real images but roughly three quarters at 9{,}000. The increase of this ratio with budget is significant on every metric, as is the encoder~$\times$~log-budget interaction (Appendix~\ref{app:stats_encoder_budget}). Given enough target-subject data the encoder can be fit without a cross-subject prior; below that it cannot. The scratch encoder still improves the decoder at every budget, and at the full budget the pre-trained encoder's advantage is no longer significant on PixCorr, Inception, CLIP or EffNet, so pre-training sharpens the effect rather than causing it. The negligible effect of general transfer in Table~\ref{tab:encoder-variants}, measured at the full budget, would not hold at small budgets.

\subsection{Additional Results for Using Synthetic Data Across Data Scales}
\label{app:few-shot-additional}
\begin{figure}[t]
  \centering
  \includegraphics[width=\linewidth]{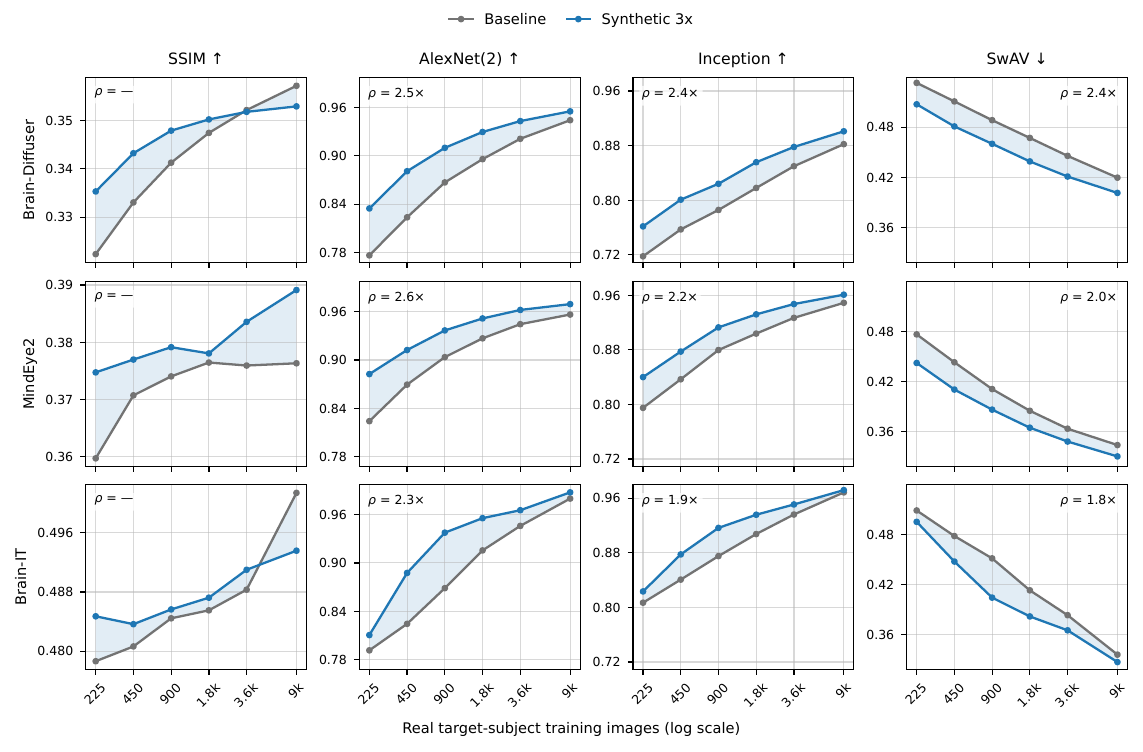}
  \caption{\textbf{Effect of synthetic data across data scales: remaining metrics.} Companion to Figure~\ref{fig:various_data_scale}, showing SSIM, AlexNet(2), Inception and SwAV for each decoder. }
\label{fig:various_data_scale_appendix}
\end{figure}

Figure~\ref{fig:various_data_scale_appendix} shows results for additional metrics that were not shown in figure~\ref{fig:various_data_scale}. While AlexNet(2), Inception, and SwAV behave similarly, SSIM is a clear outlier. $\rho$ is not reported for SSIM because its nearly flat baseline curve makes it unidentifiable for all three models (Appendix~\ref{app:stats_scaling}). It is observed that SSIM also decreases for the synthetic data model at the full 9{,}000-image budget for Brain-Diffuser and Brain-IT.

\subsection{Scaling the Amount of Synthetic Data}
\label{app:scaling-synthetic}

\begin{figure}[t]
  \centering
  \includegraphics[width=\linewidth]{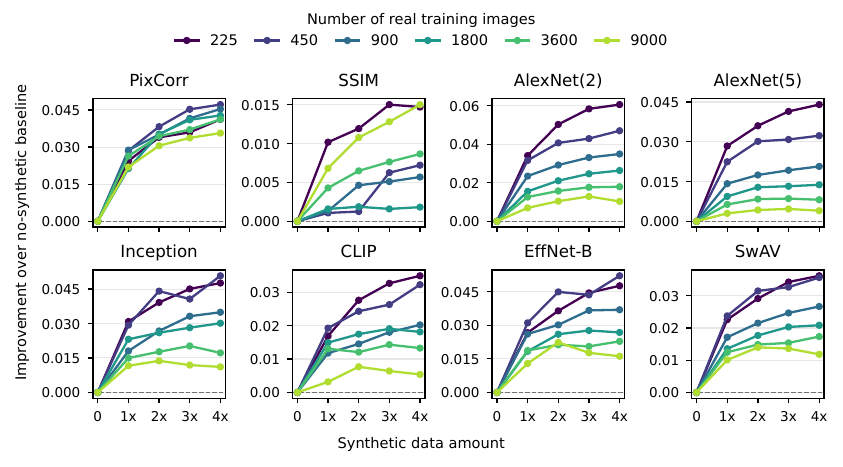}
  \caption{\textbf{Scaling the amount of synthetic data.} Improvement over the baseline for MindEye2, as a function of the number of synthetic image-fMRI pairs added per real training image. One line per real-data budget $x$; each curve is measured against its own $x$-matched $m{=}0$ baseline, so all curves start at zero. Lower-is-better metrics are sign-flipped, so \emph{up is better in every panel}; note that vertical scales differ across panels. Gains are positive at every budget, and grow as the real-data budget shrinks.}

  \label{fig:scaling_synthetic_delta}
\end{figure}

The experiments so far add a fixed amount of synthetic data. Here we vary it systematically, to establish whether the benefit is specific to one operating point and how it interacts with the amount of real data available for the target subject. Similar to Sec.~\ref{sec:scaling-synthetic}, we fine-tune MindEye2 on $x \in \{225, 450, 900, 1800, 3600, 9000\}$ real images from the target subject, where $x = 9000$ corresponds to the full budget, and additionally vary the number of images that are used for synthetic data, $y = m \cdot x$ for $m \in \{0, 1, 2, 3, 4\}$. The synthetic pairs are generated in the same manner, and the $m = 0$ condition is the real-data-only baseline. Every configuration is trained separately for subjects 1, 2, 5 and 7, and all results are averaged over those four subjects. Figure~\ref{fig:scaling_synthetic_delta} reports each metric as the improvement over that budget's own $m = 0$ baseline, so that all curves start at zero and lower-is-better metrics are sign-flipped.

\paragraph{Most of the benefit arrives by $2\times$.}
The gain is strongly concave in $m$. Averaged over the eight metrics, the
first increment ($m = 0 \to 1$) delivers 63\% (95\% CI 60-66) of the total
improvement measured at $m=4$, and $m=2$ delivers 84\% (81-87). The
remaining increments are small but not uniformly null: the $m = 1 \to 2$
step is significant on 32 of 48 cells and the $m = 2 \to 4$ step on 25 of
48, so a larger pool continues to help modestly; the only significant
reversal is EffNet at the full budget from $m=2$ to $m=4$.
 
\paragraph{The benefit grows as real data becomes scarce.}
At $m=4$, the improvement in AlexNet(2) falls monotonically from 6.1 points
at $x=225$ to 1.0 points at $x=9000$, and AlexNet(5), Inception, CLIP,
EffNet-B and SwAV show the same ordering without crossings; for all six of
these metrics the $x=225$ gain significantly exceeds the $x=9000$ gain
(Spearman $\rho \le -0.94$ between budget and gain, all $p_{\mathrm{Holm}} \le 0.001$).

\paragraph{Low-level and high-level metrics behave differently.}
The budget ordering above is clean for the semantic and identification
metrics but not for PixCorr and SSIM, where the curves are compressed and
the ordering is scrambled: at $m=4$, PixCorr improves by 0.047 at $x=450$
against 0.041 at $x=225$ and 0.035 at $x=9000$, and neither metric shows a
significant difference between the $x=225$ and $x=9000$ gains
(PixCorr $p=0.17$, SSIM $p=0.59$).

\section{Extension to Video Decoding}
\label{app:video}

We extend the \emph{encoder-generated data for decoder} recipe to fMRI-to-video decoding to test if it could be extended to other modalities.

\paragraph{Setup.}
We trained and evaluated the NEURONS ~\citep{wang2025neurons} model in the cc2017 dataset ~\citep{wen2018neural} under the limited data setting (10\% of full data) as well as the full data setting.

The cc2017 dataset comprises three subjects who viewed 18 8-minute training clips and 5 8-minute test clips inside a 3-T scanner at a 2s temporal resolution. Each fMRI sample therefore corresponds to a 2s, 6-frame video clip, giving 4320 unique training clips presented twice and 1200 test clips presented ten times; following common practice, the test repetitions are averaged.

Following the convention of Sec.~\ref{sec:scaling-synthetic}, we parameterize the
limited-data budget by the number of unique clips and retain both
repetitions of each. The target subject's training set is reduced to 432
unique clips, 10\% of the full 4,320, corresponding to 864 recorded trials
and roughly 29 minutes of scanning. The fMRI recorded for the remaining
3,888 clips is discarded; the video itself is kept.

\paragraph{Encoder adaptation and synthesis.}
The Universal Encoder~\citep{beliy2024wisdom} is jointly pre-trained on the eight NSD subjects and the two non-target cc2017 subjects, and then adapted to the target subject. This joint pre-training spans two datasets, two field strengths and two stimulus modalities at once, which the encoder accommodates without architectural change: all weights other than the per-voxel embeddings are shared across voxels, subjects and datasets, so a new subject is absorbed simply by adding its voxels to the embedding table. Because the encoder predicts responses to still images, we sample a single frame from each 6-frame clip and treat it as the stimulus for the corresponding fMRI sample. Given the synthesis pool $\mathcal{U}$, we predict a response for each 2s window independently, from its single frame. Following standard practice in video decoding, we apply a fixed temporal offset between the video and the fMRI scans, using the same offset as NEURONS when generating synthetic responses.

For the limited data experiment, we replace all of the 3,888 discarded clips with synthetic data, while for the full data experiment we introduce additional video clips from the WebVid dataset~\citep{bain2021frozen}. We use 1,200 8s clips from the dataset, using the same set of video clips selected in ~\citet{li2024enhancing}, which we split into 4,800 2s clips and generate synthetic fMRI based on them.
As in the image experiments, real and synthetic pairs are mixed within each batch and are not distinguished by the decoder in any way.

\paragraph{Evaluation metrics.}
We report the standard cc2017 metrics~\citep{wang2025neurons,gong2024neuroclips}. \emph{Video-based} metrics evaluate the reconstruction as a clip: semantic-level 2-way and 50-way top-1 accuracy from a classifier based on VideoMAE~\citep{tong2022videomae} over the 400 Kinetics-400 classes~\citep{kay2017kinetics}, and spatiotemporal-level CLIP-pcc, the average cosine similarity between CLIP embeddings of adjacent frames, which measures temporal consistency. \emph{Frame-based} metrics evaluate individual frames: semantic-level 2-way and 50-way top-1 accuracy from an ImageNet classifier, and pixel-level SSIM and PSNR.

\begin{table}[t]
\centering
\caption{\textbf{Extension to video decoding}. 
A NEURONS model is trained with and without synthetic fMRI, for limited and full data settings.
\textbf{Bold} marks the better of the two conditions for each budget.
}
\label{tab:video}
\small
\setlength{\tabcolsep}{4pt}
\begin{tabular}{llccccccc}
\toprule
& & \multicolumn{3}{c}{Video-based} & \multicolumn{4}{c}{Frame-based} \\
\cmidrule(lr){3-5}\cmidrule(lr){6-9}
& & \multicolumn{2}{c}{Semantic-level} & ST-level & \multicolumn{2}{c}{Semantic-level} & \multicolumn{2}{c}{Pixel-level} \\
\cmidrule(lr){3-4}\cmidrule(lr){5-5}\cmidrule(lr){6-7}\cmidrule(lr){8-9}
Training data & Syn. & 2-way $\uparrow$ & 50-way $\uparrow$ & CLIP-pcc $\uparrow$ & 2-way $\uparrow$ & 50-way $\uparrow$ & SSIM $\uparrow$ & PSNR $\uparrow$ \\
\midrule
\multirow{2}{*}{Limited ($432$ clips)} & \ding{55} & 0.837 & 0.185 & 0.932 & 0.770 & 0.154 & 0.357 & 9.458 \\
 & \ding{51} & \textbf{0.855} & \textbf{0.235} & \textbf{0.932} & \textbf{0.797} & \textbf{0.187} & \textbf{0.383} & \textbf{9.669} \\
\midrule
\multirow{2}{*}{Full ($4{,}320$ clips)} & \ding{55} & 0.857 & 0.259 & \textbf{0.937 }& 0.810 & 0.206 & 0.377 & 9.672 \\
& \ding{51} & \textbf{0.865} & \textbf{0.275} & 0.936 & \textbf{0.816} & \textbf{0.221} & \textbf{0.397} & \textbf{9.896 }\\
\bottomrule
\end{tabular}
\end{table}

\paragraph{Results.}
Table~\ref{tab:video} shows that encoder-generated fMRI recovers most of the
performance lost to the data limitation. Under the limited budget, synthetic
data significantly improves every metric except CLIP-pcc, which the data
limitation itself barely affects (a drop of 0.005). The limited-data model
with synthetic data recovers 63-87\% of the gap to the full-data model on
the four semantic metrics and matches or exceeds it on the pixel-level
metrics (SSIM 127\% [107, 153], PSNR 99\% [71, 139]). At the full budget the
gains are smaller but remain significant on five of seven metrics
(Appendix~\ref{app:stats_video}).

Two aspects of this result are worth emphasizing. First, the encoder is a still-image model: it never observes motion, and each synthetic response is predicted from one frame of a six-frame clip. Such coarse supervision still improves a video decoder.
Second, no component of either model was modified for the video case. This supports the claim of Sec.~\ref{sec:synthetic_recipe} that the intervention is at the level of the data rather than the model, and suggests the approach may carry to other stimulus and recording modalities for which a stimulus-to-signal encoder is available.
We intend this experiment as a preliminary check and deliberately kept the setup simple: a still-image encoder, a single frame per 2s window, and no temporal modeling. Whether an encoder that models temporal dynamics would improve the value of synthetic data remains to be tested.

\section{Statistical Tests}
\label{app:stats}

This section describes the statistical procedures behind the claims in the main text and the appendix, and summarizes their outcomes. Point estimates are given in the corresponding result tables and are not repeated here; per-contrast estimates, confidence intervals and $p$-values not listed below are provided with the released results. Unless stated otherwise, tests are two-sided, $p$-values are Holm-corrected within the stated family, and confidence intervals (CIs) are 95\% percentile bootstrap intervals from 10{,}000 resamples. Since each condition is trained once, all tests account for the sampling of subjects, concepts and test images, but not for variation across training runs.

\subsection{Concept generalization across decoders}
\label{app:stats_decoders}

\paragraph{Procedure.}
For each decoder and metric, we average within each of the 48 subject-concept cells (4 subjects $\times$ 12 supercategories) and compare conditions with Wilcoxon signed-rank tests over cells, Holm-corrected across the 20 planned comparisons per decoder (10 metrics $\times$ 2 contrasts). CIs are bootstrapped over cells, stratified by subject, and recovery fractions are computed from aggregate means. As sensitivity checks, we repeat each comparison with paired $t$-tests and with linear mixed models with crossed random intercepts for subject and concept, and, for the two recall metrics, with logistic mixed models on individual qualifying images. The two-way identification metrics are computed against the full pool of remaining test images, so no distractor sampling enters the comparison. The complementary test set (Appendix~\ref{app:complementary_results}) is analyzed in the same way, with 16 comparisons per decoder (8 metrics $\times$ 2 contrasts).

\paragraph{Results.}
Both the concept generalization gap (Seen vs.\ Unseen) and the gain from synthetic data (Unseen + syn.\ vs.\ Unseen) are significant for every decoder on every metric except SSIM, which the holdout leaves essentially unchanged for Brain-Diffuser and Brain-IT and shifts by only 0.004 for MindEye2. All sensitivity analyses agree, with one exception: the PixCorr gain for Brain-IT is significant under the Wilcoxon test ($p_{\text{Holm}} = 0.022$) but not under the mixed model ($p = 0.063$). Object Recall improves for each of the twelve concepts under every decoder. Table~\ref{tab:stats_recovery} reports the fraction of the gap recovered by synthetic data, and Table~\ref{tab:stats_complementary} compares the effect of each manipulation on images with and without \concept.

\begin{table}[t]
\centering\small
\caption{Fraction of the concept generalization gap recovered by synthetic data, $(\text{syn} - \text{unseen})/(\text{seen} - \text{unseen})$ on aggregate means, with 95\% bootstrap CIs over the 48 subject-supercategory cells. Left: the three decoders of Table~\ref{tab:unseen-concept-decoder}. Right: the probe decoder of Table~\ref{tab:encoder-variants} under each synthesis encoder. SSIM is omitted because the holdout barely changes it. Subject-level bootstrap intervals are similar and are provided with the released results.}
\label{tab:stats_recovery}
\resizebox{\linewidth}{!}{
\begin{tabular}{l ccc ccc}
\toprule
& \multicolumn{3}{c}{Decoders (Table~\ref{tab:unseen-concept-decoder})} & \multicolumn{3}{c}{Probe decoder by encoder (Table~\ref{tab:encoder-variants})} \\
\cmidrule(lr){2-4}\cmidrule(lr){5-7}
Metric & Brain-Diffuser & MindEye2 & Brain-IT & Scratch-blind & Transfer-blind & Transfer-concept \\
\midrule
PixCorr & 82\% [77, 87] & 71\% [62, 81] & 40\% [16, 66] & 61\% [51, 74] & 59\% [49, 72] & 73\% [63, 86] \\
Alex(2) & 92\% [84, 102] & 64\% [54, 72] & 64\% [38, 89] & 63\% [51, 76] & 67\% [57, 79] & 77\% [67, 88] \\
Alex(5) & 93\% [89, 98] & 73\% [68, 79] & 58\% [46, 71] & 63\% [56, 72] & 64\% [57, 73] & 74\% [67, 81] \\
Incep & 83\% [79, 87] & 66\% [60, 71] & 46\% [33, 55] & 51\% [46, 56] & 51\% [46, 57] & 69\% [63, 74] \\
CLIP & 87\% [84, 90] & 57\% [53, 62] & 53\% [43, 62] & 46\% [41, 53] & 49\% [43, 56] & 62\% [57, 67] \\
EffNet & 76\% [74, 79] & 53\% [49, 58] & 38\% [30, 44] & 37\% [33, 41] & 41\% [37, 46] & 52\% [49, 56] \\
SwAV & 81\% [79, 83] & 56\% [51, 61] & 37\% [28, 44] & 39\% [35, 43] & 43\% [39, 48] & 52\% [48, 56] \\
Object Recall & 78\% [70, 85] & 55\% [50, 60] & 41\% [32, 48] & 41\% [36, 45] & 44\% [39, 49] & 59\% [54, 64] \\
Relaxed Recall & 87\% [83, 90] & 57\% [52, 62] & 41\% [32, 49] & 48\% [44, 52] & 51\% [47, 57] & 62\% [58, 67] \\
\bottomrule
\end{tabular}
}
\end{table}

\begin{table}[t]
\centering\small
\caption{Effect of the training-set manipulations on images that do \emph{not} contain \concept, compared with their effect on images that do. $\Delta_{\mathcal{C}}$ and $\Delta_{\neg\mathcal{C}}$ are mean paired differences over the 48 subject-concept cells on the concept test set (Table~\ref{tab:unseen-concept-decoder}) and the complementary test set (Table~\ref{tab:complementary-decoder}); positive = improvement, EffNet and SwAV sign-flipped, identification metrics in percentage points. Ratio is $\Delta_{\neg\mathcal{C}}/\Delta_{\mathcal{C}}$. $^{*}$ marks $\Delta_{\neg\mathcal{C}}$ significant after Holm correction (Wilcoxon signed-rank, 16 comparisons per decoder).}
\label{tab:stats_complementary}
\begin{tabular}{ll rrr rrr}
\toprule
& & \multicolumn{3}{c}{Seen vs.\ Unseen} & \multicolumn{3}{c}{Unseen + syn.\ vs.\ Unseen} \\
\cmidrule(lr){3-5}\cmidrule(lr){6-8}
Decoder & Metric & $\Delta_{\mathcal{C}}$ & $\Delta_{\neg\mathcal{C}}$ & Ratio & $\Delta_{\mathcal{C}}$ & $\Delta_{\neg\mathcal{C}}$ & Ratio \\
\midrule
\multirow{4}{*}{Brain-Diffuser} & Incep & 9.0 & 0.4$^{*}$ & 4\% & 7.5 & 0.2 & 3\% \\
 & CLIP & 9.3 & 0.1 & 1\% & 8.1 & 0.3$^{*}$ & 4\% \\
 & EffNet & 0.086 & 0.001 & 1\% & 0.066 & 0.003$^{*}$ & 4\% \\
 & SwAV & 0.061 & 0.002$^{*}$ & 4\% & 0.050 & 0.004$^{*}$ & 8\% \\
\midrule
\multirow{4}{*}{MindEye2} & Incep & 7.4 & 0.3$^{*}$ & 5\% & 4.9 & 0.4$^{*}$ & 8\% \\
 & CLIP & 9.5 & 0.8$^{*}$ & 8\% & 5.5 & 0.4$^{*}$ & 8\% \\
 & EffNet & 0.132 & 0.009$^{*}$ & 7\% & 0.070 & 0.010$^{*}$ & 14\% \\
 & SwAV & 0.070 & 0.005$^{*}$ & 8\% & 0.040 & 0.006$^{*}$ & 14\% \\
\midrule
\multirow{4}{*}{Brain-IT} & Incep & 2.5 & 0.2$^{*}$ & 7\% & 1.1 & 0.2 & 15\% \\
 & CLIP & 3.0 & 0.0 & 1\% & 1.6 & 0.0 & 0\% \\
 & EffNet & 0.059 & 0.002 & 3\% & 0.022 & 0.000 & -2\% \\
 & SwAV & 0.029 & 0.001 & 2\% & 0.011 & 0.000 & 4\% \\
\bottomrule
\end{tabular}
\end{table}

\subsection{Encoder variants}
\label{app:stats-encoder}

\paragraph{Procedure.}
For Tables~\ref{tab:encoder-variants} and~\ref{tab:encoder-variants-category}, we use the cell-level paired procedure of Appendix~\ref{app:stats_decoders} over 48 subject-supercategory or 88 subject-category cells, Holm-corrected within each granularity and family of tests. Equivalence is assessed with two one-sided tests (TOST) at a margin of 25\% of the relevant oracle-to-blind gap: the unseen-to-real gap for probe-decoder contrasts, and the transfer-oracle to transfer-blind gap for encoder contrasts. Whether a contrast depends on granularity is tested with a mixed model that includes a granularity $\times$ condition term and crossed random intercepts for subject and concept. Because no transfer variant sees \concept in the target subject while withholding it from the other subjects, the two-factor design is incomplete, and we test the estimable contrasts directly. Agreement between an encoder-side metric and decoder benefit is the Spearman correlation, within each cell, between that metric and probe-decoder Object Recall across the three synthesis variants, averaged over cells. For decoded recall, we additionally fit image-level logistic mixed models with random intercepts for subject and concept. Point estimates for every contrast below are differences between rows of Tables~\ref{tab:encoder-variants} and~\ref{tab:encoder-variants-category}.

\paragraph{Probe decoder.}
Transfer-blind and scratch-blind are statistically equivalent on nine of ten probe-decoder metrics at supercategory granularity and on five of ten at category granularity, so general transfer adds little at the full budget. Concept-specific transfer (transfer-concept vs.\ transfer-blind) is significant on every metric at supercategory granularity and on none at category granularity, and this dependence on granularity is itself significant for Inception, CLIP and EffNet.

\paragraph{Fidelity.}
Removing \concept from all subjects lowers encoder Pearson $r$ significantly at both granularities, but pre-training on other subjects raises it by more (by a further 0.020 [0.012, 0.026] at supercategory and 0.039 [0.035, 0.043] at category granularity; all $p < 0.001$). The within-cell rank correlation between $r$ and probe-decoder Object Recall is $\rho = 0.56$ [0.42, 0.70] at supercategory granularity and 0.10 [$-0.04$, 0.24] at category granularity.

\paragraph{Decoded recall.}
Removing \concept from the encoder's training data, from all subjects or from the target subject only, significantly lowers decoded Object Recall and Relaxed Recall at both granularities ($p_{\text{Holm}} < 10^{-4}$, image-level models agree). Removing it from all subjects costs 34\% [30, 38] and 20\% [17, 25] of the oracle's decoded Object Recall and Relaxed Recall at supercategory granularity, and 18\% [14, 22] and 6\% [4, 8] at category granularity.

Decoded recall orders pre-training and exposure opposite to fidelity. At supercategory granularity, scratch-oracle significantly exceeds transfer-blind on both decoded metrics ($p_{\text{Holm}} < 10^{-5}$, image-level models agree), and transfer-concept is equivalent to scratch-oracle on both. At category granularity, the Object Recall reversal is not robust (significant by Wilcoxon, $p_{\text{Holm}} = 0.003$, but not under the mixed model, $p = 0.59$), Relaxed Recall does not differ, and transfer-concept and scratch-oracle are equivalent on Object Recall only. For supercategory Relaxed Recall, the cell-level difference between transfer-concept and scratch-oracle is null ($p = 0.92$) while the image-level model detects one ($p = 0.004$), which reflects weighting by image rather than by cell.

Unlike the probe decoder, decoded recall credits both transfer steps. At supercategory granularity, general transfer (transfer-blind vs.\ scratch-blind) and concept-specific transfer (transfer-concept vs.\ transfer-blind) are both significant on both metrics ($p_{\text{Holm}} < 10^{-6}$) and do not differ significantly from each other ($p_{\text{Holm}} = 0.65$ for both metrics). At category granularity, general transfer is the larger step ($p_{\text{Holm}} = 0.021$ for Object Recall and 0.027 for Relaxed Recall, the latter not significant under the mixed model). Transfer-blind and scratch-blind are not equivalent on decoded recall at either granularity (TOST $p = 1.00$). Within cells, decoded recall agrees with probe-decoder Object Recall about as well as $r$ does ($\rho = 0.58$ and 0.55 for decoded Object Recall and Relaxed Recall at supercategory granularity, 0.14 and 0.07 at category granularity).

\paragraph{Encoder versus decoder retention (Fig.~\ref{fig:retained}).}
Per cell, encoder retention is transfer-concept over transfer-oracle, and decoder retention is MindEye2 unseen over seen, computed on the 48 supercategory cells (no denominator is zero). Aggregate encoder retention is 0.869 [0.847, 0.889] for Object Recall and 0.931 [0.917, 0.946] for Relaxed Recall, against 0.216 [0.175, 0.260] and 0.365 [0.304, 0.427] for the decoder. Encoder retention exceeds decoder retention in every cell, on average by 0.638 [0.587, 0.688] and 0.561 [0.505, 0.617] ($p_{\text{Holm}} < 10^{-13}$); subject-level bootstrap intervals are similar.

\paragraph{Sensitivity.}
Paired $t$-tests and linear mixed models agree with the Wilcoxon conclusions on every contrast except three: the category-level Object Recall reversal above, the category-level difference in $r$ between scratch-oracle and scratch-blind, and the category-level difference in Relaxed Recall between the two transfer steps. Excluding the four subcategories with fewer than five qualifying images (toaster, parking meter, hair drier, toothbrush) from Object Recall changes no decision.

\subsection{Synthetic-data controls and frozen backbone}
\label{app:stats_frozen}

\paragraph{Procedure.}
All models in Table~\ref{tab:controls_frozen} are trained on subject~1 and evaluated on the same test images, so we use a paired stratified bootstrap over images. Images are resampled within each subcategory's qualifying set $Q_o$ for Object Recall, within each supercategory's qualifying set $Q_\gamma$ for Relaxed Recall, and within each supercategory's test set for the remaining metrics. The same resampled indices are used for every condition, and stratum sizes and macro-averaging weights are held fixed. Each control is tested against the unseen baseline on all ten metrics (20 comparisons) and against transfer-concept on the two recall metrics (4 comparisons). Equivalence, of the non-\concept control to the unseen baseline and of each frozen encoder to its LoRA counterpart, is assessed with TOST at the margin of Appendix~\ref{app:stats-encoder}, reporting 90\% CIs. SSIM is excluded from equivalence claims because the holdout barely changes it, which makes its margin degenerate.

\paragraph{Results.}
Both controls fall significantly below transfer-concept on both recall metrics ($p_{\text{Holm}} < 0.001$). The non-\concept control is equivalent to the unseen baseline on all four high-level metrics and both recall metrics, and significantly improves PixCorr, SSIM and Alex(5). Wrong fMRI significantly raises both recall metrics and significantly lowers Alex(2), Alex(5) and SSIM.

Both frozen encoders are equivalent to their LoRA counterparts on all four high-level metrics and both recall metrics ($p_{\text{Holm}} \le 0.002$), with 90\% CIs for the Object Recall difference within $\pm 0.033$. The low-level metrics are mostly inconclusive; the only significant difference is a small SSIM decrease for transfer-blind. With a frozen backbone, transfer-concept still significantly exceeds transfer-blind on both recall metrics ($p_{\text{Holm}} < 0.001$). Whether freezing reduces this advantage is unresolved: the difference in advantage is 0.030 [0.000, 0.059] for Object Recall and 0.019 [$-0.020$, 0.058] for Relaxed Recall, and at a stricter margin of half the LoRA advantage, equivalence holds only for Relaxed Recall of transfer-concept. Excluding the four subcategories with fewer than five qualifying images changes none of these decisions.

\subsection{Data scaling}
\label{app:stats_scaling}

\paragraph{Procedure.}
In the scaling experiments of Sec.~\ref{sec:scaling-synthetic}, the four subjects share the same 1{,}000 test images, so we compare conditions at the image level. Each of the 48 budget-metric contrasts per decoder (6 budgets $\times$ 8 metrics) is the mean paired difference between training with and without synthetic data over the 4{,}000 subject-image pairs, with CIs and $p$-values from bootstrap resamples of images within subject, Holm-corrected within decoder. With only four subjects, a subject-level sign test cannot reach $p < 0.05$, so the image-level tests are primary.

For the data-efficiency factor $\hat\rho$, the monotone interpolant $\hat b$ is a PCHIP fit to the six baseline points on $\log x$, and the interpolant and $\hat\rho$ are refitted on each bootstrap replicate. Because $\hat\rho x$ exceeds 9{,}000 at the largest budget for every metric, $\hat b$ is held at its 9{,}000-image value there rather than extrapolated. The largest budget therefore does not inform the fit; measuring the factor there would require a baseline trained on roughly twice the full budget, which cannot be obtained without collecting new data from the NSD subjects.

\paragraph{Results.}
Synthetic data significantly improves 137 of the 144 contrasts. The two significant reversals are SSIM at the full budget for Brain-Diffuser and Brain-IT (Table~\ref{tab:full_limited_results}); the five non-significant contrasts are SSIM at intermediate budgets and, for Brain-IT at the full budget, Inception and CLIP. At the subject level, all four subjects improve in 46, 47 and 41 of the 48 contrasts for Brain-Diffuser, MindEye2 and Brain-IT. Every curve still rises significantly from 3{,}600 to 9{,}000 real images, except SSIM for the MindEye2 baseline. Table~\ref{tab:stats_rho} lists $\hat\rho$ with image-level CIs; subject-level intervals are wider but overlap them for every entry. The improvement from synthetic data decreases monotonically with budget on five of eight metrics for Brain-Diffuser and MindEye2, and on none for Brain-IT, whose gain is non-monotonic across budgets.

\begin{table}[t]
\centering\small
\caption{Data-efficiency factor $\hat\rho$ for every metric and decoder, with 95\% CIs from 10,000 subject-stratified image-level bootstraps in which the interpolant and $\hat\rho$ are refitted on each replicate. SSIM is omitted: its curves are nearly flat across budgets, so $\hat\rho$ is not identifiable for it (point estimates of 0.8 to 4.9 with CIs spanning an order of magnitude).}
\label{tab:stats_rho}
\begin{tabular}{l ccc}
\toprule
Metric & Brain-Diffuser & MindEye2 & Brain-IT \\
\midrule
PixCorr & 2.64 [2.48, 2.77] & 2.94 [2.67, 3.26] & 1.94 [1.87, 2.02] \\
Alex(2) & 2.55 [2.42, 2.69] & 2.61 [2.45, 2.80] & 2.27 [2.17, 2.38] \\
Alex(5) & 2.39 [2.27, 2.53] & 2.35 [2.20, 2.51] & 2.03 [1.91, 2.19] \\
Incep & 2.37 [2.20, 2.64] & 2.16 [1.98, 2.36] & 1.86 [1.72, 1.99] \\
CLIP & 2.20 [2.08, 2.35] & 1.74 [1.61, 1.87] & 1.78 [1.66, 1.90] \\
EffNet & 2.30 [2.21, 2.40] & 1.87 [1.78, 1.96] & 1.78 [1.72, 1.83] \\
SwAV & 2.43 [2.35, 2.52] & 1.99 [1.91, 2.08] & 1.84 [1.79, 1.89] \\
\bottomrule
\end{tabular}
\end{table}

\subsection{Encoder pre-training across budgets}
\label{app:stats_encoder_budget}

\paragraph{Procedure.}
Figure~\ref{fig:encoder-pretrain-scales} compares three conditions per budget: the baseline and synthetic data from either the pre-trained or the scratch encoder. We use the image-level paired bootstrap of Appendix~\ref{app:stats_scaling}, Holm-corrected within metric. The budget dependence of the scratch encoder's relative benefit is tested in two ways: a one-sided paired bootstrap of the change, from 225 to 9{,}000 images, in the ratio of scratch to pre-trained improvement, and a mixed model on subject-level scores with an encoder $\times$ $\log x$ term and a random intercept per subject.

\paragraph{Results.}
Both encoders improve on the baseline in nearly every budget-metric contrast (47 of 48 for the pre-trained encoder and 42 of 48 for the scratch encoder; the exceptions are SSIM and, for the scratch encoder, CLIP at three budgets). The pre-trained encoder's advantage is significant in 41 of 48 contrasts; it is not significant for SSIM at 225-900 images, nor at 9{,}000 images for PixCorr, Inception, CLIP and EffNet. Averaged over the four metrics in the upper panel of Figure~\ref{fig:encoder-pretrain-scales}, the ratio of scratch to pre-trained improvement is 0.21 at 225 images and 0.77 at 9{,}000. Its increase with budget is significant on all eight metrics, and the mixed model confirms the encoder $\times$ budget interaction on all eight ($p_{\text{Holm}} \le 0.04$).

\subsection{Scaling the amount of synthetic data}
\label{app:stats_synth_scaling}

\paragraph{Procedure.}
For Figure~\ref{fig:scaling_synthetic_delta}, we use the image-level paired bootstrap of Appendix~\ref{app:stats_scaling}, comparing each multiplier $m$ with both the $x$-matched $m = 0$ baseline and the preceding multiplier, Holm-corrected within metric. The fraction of the $m = 4$ gain reached at $m = 1$ and $m = 2$ is computed per metric as the ratio of gains averaged over the six budgets, then averaged over metrics, with a bootstrap CI. The dependence of the $m = 4$ gain on budget is tested with a bootstrap Spearman correlation between $x$ and the gain, and with a one-sided paired bootstrap of the $x = 225$ gain minus the $x = 9{,}000$ gain.

\paragraph{Results.}
Both budget tests are significant for AlexNet(2), AlexNet(5), Inception, CLIP, EffNet and SwAV ($p_{\text{Holm}} \le 0.001$), and neither is significant for PixCorr or SSIM.

\subsection{Video decoding}
\label{app:stats_video}

\paragraph{Procedure.}
The three cc2017 subjects share the same 1{,}200 test clips, so we compare training with and without synthetic data at the clip level ($n = 3{,}600$ subject-clip pairs), using Wilcoxon signed-rank tests Holm-corrected within budget and CIs from subject-stratified clip bootstraps. Paired $t$-tests agree throughout. The identification metrics use one fixed distractor assignment, shared by all four conditions of Table~\ref{tab:video}.

\paragraph{Results.}
Under the limited budget, synthetic data significantly improves six of seven metrics ($p_{\text{Holm}} < 10^{-6}$), leaving only CLIP-pcc unchanged. Under the full budget, it significantly improves five of seven; Frame 2-way and CLIP-pcc do not reach significance (both $p = 0.06$). All three subjects improve individually on every metric except CLIP-pcc at both budgets and Frame 50-way at the full budget. The fractions of the limited-to-full gap recovered by synthetic data are 87\% [56, 127] for Video 2-way, 68\% [53, 85] for Video 50-way, 69\% [57, 82] for Frame 2-way, 63\% [48, 80] for Frame 50-way, 127\% [107, 153] for SSIM and 99\% [71, 139] for PSNR; CLIP-pcc is excluded because its gap is negligible.

\end{document}